\documentclass[runningheads]{llncs}

\usepackage{eccv}

\usepackage{eccvabbrv}

\usepackage{graphicx}
\usepackage{booktabs}

\usepackage[accsupp]{axessibility}  

\usepackage{hyperref}
\usepackage{orcidlink}

\usepackage{url}
\usepackage{graphicx}
\usepackage{xcolor}
\usepackage[dvipsnames]{xcolor}
\usepackage{booktabs}
\usepackage{multicol}
\usepackage{multirow}
\usepackage{amsmath}
\usepackage{epigraph} 
\usepackage{pdfpages}
\usepackage{subcaption}
\definecolor{first_level_green}{HTML}{D9F2D0}
\definecolor{second_level_blue}{HTML}{C1E3F0}
\definecolor{third_level_purple}{HTML}{F2CFEE}
\definecolor{inference_blue}{HTML}{DCEAF7}
\definecolor{loser_red}{HTML}{C00200}
\definecolor{winner_green}{HTML}{8ED973}
\definecolor{light_Apricot}{HTML}{FDE6D6}

\title{Sense it with your eyes: Sensation Generation and Understanding for Advertisements}

\author{Aysan Aghazadeh \and
Sina Malakouti \and
Adriana Kovashka}
\authorrunning{}
\institute{University of Pittsburgh}

\begin{document}

\maketitle

\begin{abstract}

Sensory advertising evokes human senses through visual cues, enabling audiences to mentally simulate experiences and increasing persuasive impact. Despite the recent increase in using AI in generating and understanding creative and persuasive content, how advertisements visually evoke sensations remains largely unexplored. In this work, we introduce the first study of understanding, evaluating, and generating sensory ads. We introduce the Sensory Ad dataset, and define sensation classification tasks (SenseClass) to benchmark LLMs and MLLMs. We further propose SenseScore, an automated evaluation metric for sensation evocation achieving strong agreement with human judgments. Finally, we introduce the Sensory Ad Generation (SenseGen) task and propose SAGA, a multi-agent framework that improves message–image alignment, sensory evocation, and persuasion. Our work establishes a foundation for sensory-aware visual persuasion.\\ Code is available at \href{https://github.com/aysanaghazadeh/SensoryAds}{https://github.com/aysanaghazadeh/SensoryAds}
\end{abstract}

\section{Introduction}

\epigraph{{\scriptsize ``I have left behind illusion, I said to myself. Henceforth I live in a world of three dimensions--with the \textbf{aid of my five senses}. I have since learned that there is no such world, but then, as the car turned out of sight of the house, I thought it took no finding, but lay all about me at the end of the avenue.'' (Evelyn Waugh, ``Brideshead Revisited'')}}

\begin{figure}[!t]
    \centering
    \includegraphics[width=0.9\linewidth]{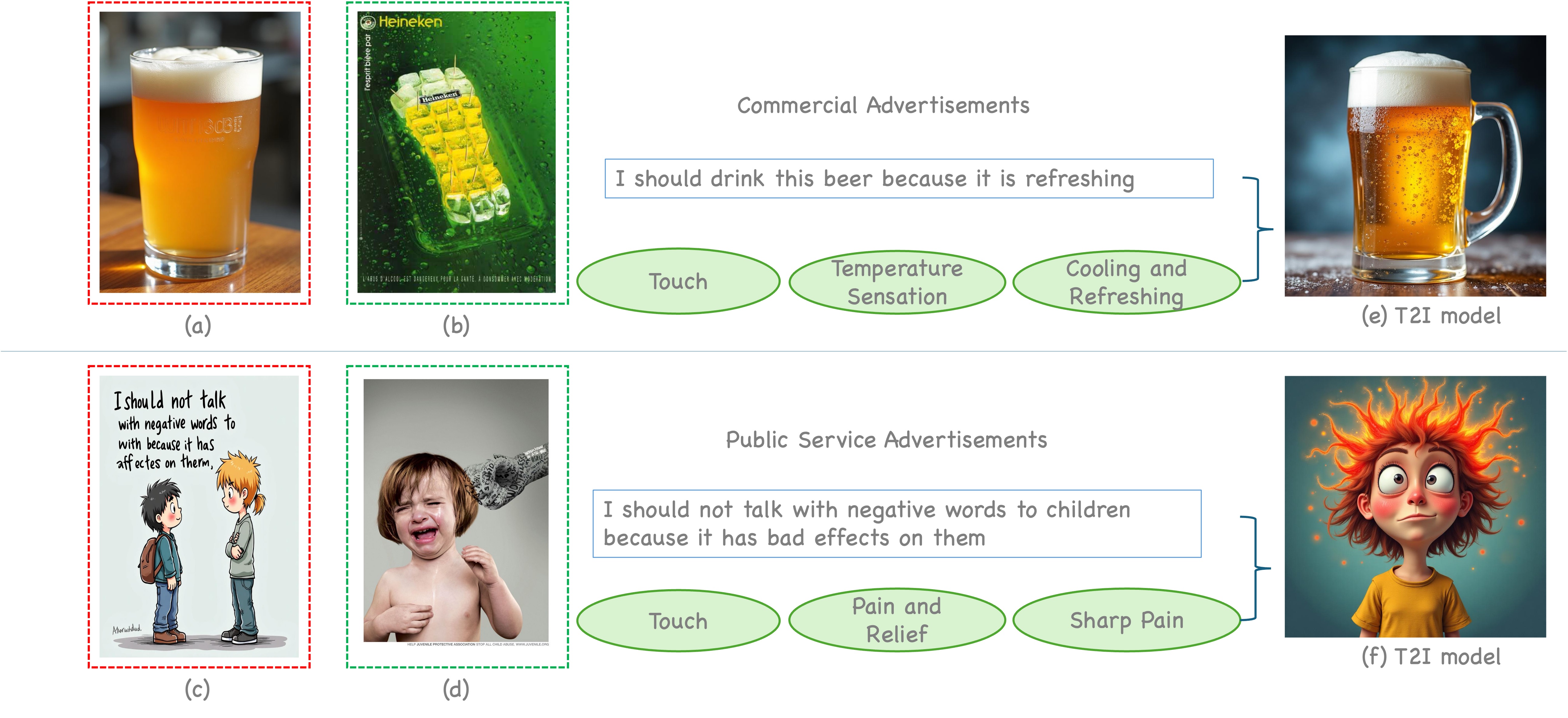}
    \caption{\textbf{Sensory Ad Generation:} We show two rows containing intended ad messages and possible images for these messages. The message text is shown in a box (``I should... because...''). The first column of images (a, c) are images generated by a T2I model using the message as prompt. The images in the second column (b, d) are real human-designed ads. The third column (e, f) are ads designed by a T2I model which also takes an intended sensation as input.} 
    \label{fig:intro}
\end{figure}

The full spectrum of senses is important for humans to navigate and experience their environments. However, humans sometimes hallucinate sensations, with very real effects: people experiencing lexical-gustatory synesthesia experience taste triggered by words \cite{ward2003lexical},
visually impaired people can ``see'' with their tongue through electrical signals \cite{nau2015acquisition}, phantom limb pain can be treated with augmented reality \cite{prahm2025phantomar}.
In marketing, sensory advertising 
enhances persuasiveness and brand effectiveness 
\cite{krishna2012integrative,lindstrom2006brand,krishna2016power,elder2022review} by allowing the audience to imagine the benefit of a product or the consequence of an action in a visceral way. 
Since stimulating the senses in the exact sense modality is infeasible, ads resort to visual content \emph{associated} with the target sensation. For example in Fig.~\ref{fig:intro}, 
on a hot summer day, image (b) is more likely to convince a thirsty audience to buy the drink by evoking the cooling and refreshing sensation (through the inclusion of the ice cubes), compared to (a). 

In this work, we conduct the first investigation of \emph{how ads evoke the senses through visual means}. We focus on three facets: recognizing, scoring (evaluating), and generating sensory evocative ads. 
First, we develop a taxonomy of senses at different levels of granularity in which the first layer corresponds to the five fundamental sensory modalities (information perceived through the eyes, ears, nose, skin, and tongue). These senses are then further refined into more specific subcategories (e.g., ``temperature'' is a type of ``touch''). 
We construct a dataset (primarily intended for evaluation) 
by collecting annotations 
on 670 images sourced from an existing dataset of advertisements (PittAds \cite{PittAd}): 
whether the image evokes a sensation and if so, the category of sensation, the visual elements evoking it, and score of \emph{how well} the image evokes the sensation.

Second, because evoking some categories of sensations can be triggering for vulnerable groups, the identification and filtering of such content is critical. We introduce two sensation classification tasks (\textbf{SenseClass}) to evaluate the capability of LLMs and MLLMs in detecting sensory modalities within ads.

Third, we propose an evaluation method, \textbf{SenseScore}, that measures how effectively an image evokes a target sensation. SenseScore first utilizes an MLLM to generate the description of the image, then fine-tunes an LLM using Contrastive Preference Optimization \cite{CPO}. 
Experimental results show that our evaluation metric achieves a Kappa \cite{kappa} agreement score of 0.8 with human annotators, representing an improvement of 40\% over existing baseline metrics.

Fourth, we introduce the \textbf{Sensory Ad Generation (SenseGen)} task, where the goal is to generate advertisement images that both convey a given message and evoke a specified sensation. The messages are 
in the form \textit{``I should \{action\} because \{reason\}''} \cite{PittAd}.
Our results show that existing T2I models fail 
in generating advertisement images that evoke specific sensations while accurately conveying the intended message. 

Finally, we propose a multi-agent image generation framework, \textbf{Sensory Ad Generation with Agent (SAGA)}, to generate advertisement images that both convey the message and evoke the sensation. Our framework improves sensation evocation, as well as the effectiveness of the generated ads, measured via alignment of image and message, and persuasion scores from prior work \cite{CAP}.

To summarize, we introduce: (1) the \textbf{Sensory Ad dataset}, 
    (2) two sensation classification tasks which innovatively use the proposed sensation hierarchy,
    (3) the novel task of Sensory Ad Generation.
    We propose (4) \textbf{SenseScore}, an automated evaluation method for sensation evocation, and
    (5) \textbf{SAGA}, a multi-agent framework for generating effective sensory ads.

\section{Related Works}

\textbf{Text-to-Image Generation.}
Text-to-Image (T2I) models such as Flux \cite{FLUX}, Stable Diffusion \cite{SD3}, Qwen-Image \cite{QwenImage}, PixArt \cite{PixArt}, DALLE3 \cite{DALLE3}, etc. have advanced in generating high quality and realistic images given the explicit description of the prompt. Some existing work \cite{CAP, liao2024text,menon2024moodsmith} assess the capability of models in generating images from abstract concepts and messages (like advertisement design tasks). Others \cite{EmoGen, dang2025emoticrafter, park2020emotional} tackle emotion transfer through images. Sensation and emotion differ; the former is more visceral/physical while the latter is more interpretative/psychological. 
For example, in Fig. \ref{fig:intro} both image (c) and (d) can transfer sadness, but only image (d) evokes the pain sensation. 

\textbf{Text to Image Evaluation.}
Existing metrics \cite{lin2024evaluating, ImageReward} are designed to assess how well an image corresponds to an explicit prompt specifying concrete objects, attributes, or relations between visual elements. 
Evaluating \textbf{sensation evocation} poses a unique challenge: the sensation is not only an implicit concept but the same sensation can be represented through entirely different visual designs. 

\textbf{Understanding Modalities beyond Sight.}
Our work is part of a bigger trend, including examples such as understanding audio and touch data \cite{ghosh2024gama,yang2024bindingtouch} or semantic-taste and visual-smell mappings \cite{bender2023learningtastewine,ozguroglu2025smell}. Other work predicts physical properties such as density and hardness from images and descriptions \cite{zhai2024physical}. However, no prior work studies how images are created to evoke specific sensations, nor predicts computationally the impact of sensations on an audience. 

\textbf{Understanding and Generating Advertisements.}
\cite{PittAd} pioneer computational visual ad understanding, but do not capture sensory information. 
\cite{kumar2023persuasion,singh2025measuring,qiu2025mmpersuade} study persuasion strategies. 
Prior work has investigated T2I models for generating advertisements, focusing on criteria such as creativity and persuasion 
\cite{CAP}, depicting specific metaphorical relationships \cite{akula2023metaclue}, or personalized effects \cite{kim2025pvp}. However, these generation studies do not examine the models’ ability to implement specific persuasion strategies, such as the evocation of specific sensations, which play a crucial role in making ads influential and memorable. 

\textbf{Sensory Advertising} is studied in \cite{krishna2014sensory,petit2019digital,hulten2015sensory,krishna2016power}. 
Krishna et al. \cite{krishna2012integrative} define sensory marketing as ``marketing that engages the consumers' senses and affects their perception, judgment and behavior.'' Subconscious sensory triggers may make the ad's message more compelling than explicit messaging, including causing viewers to perceive specific properties of the product. 
Examples include product packaging (e.g., Hershey's chocolate kisses creating the sensation of a drop melting), sound symbolism (e.g., the word ``frosh'' evoking the sensation of creaminess more than ``frish''), and the memories scents create and evoke. 
Cian et al. \cite{cian2014logo} describe the dynamics encoded in similar but slightly varied imagery (e.g., a horizontal vs tilted seesaw).
None of this work computationally models visually evoked sensations. 
\cite{singh2024teaching} focus on 
 observable behavioral reactions such as likes, upvotes, or memorability, which are external message recipient (viewer) outcomes and do not indicate what sensory experiences an image evokes. We instead focus on the sender of the signals (the images that evoke the sensations).
\section{Dataset, Tasks and Methods}

\begin{figure}[!tp]
    \centering
    \includegraphics[width=1\linewidth]{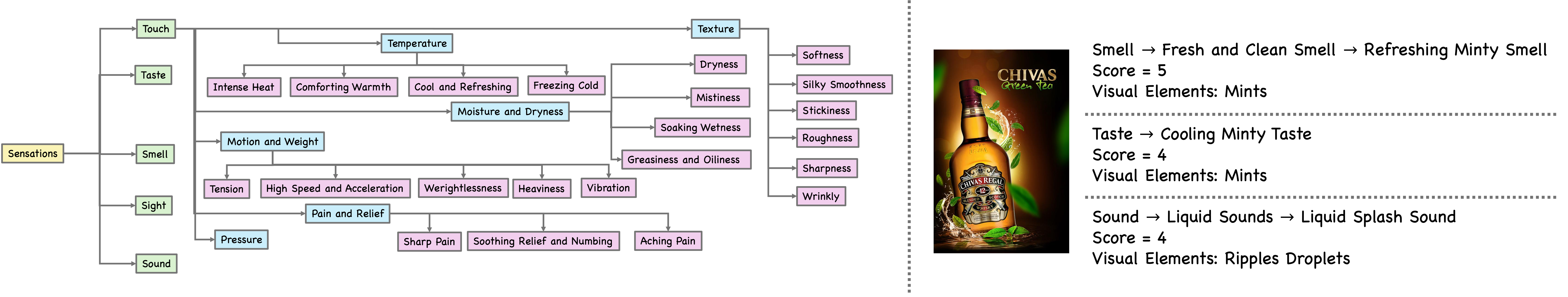}
    \vspace{-0.6cm}
    \caption{Left: \textbf{Sensation hierarchy} (only categorizing the Touch sensation): \colorbox{first_level_green}{Green box} represents first level sensation, \colorbox{second_level_blue}{blue} represents second, and \colorbox{third_level_purple}{purple} represents third.  Right: \textbf{Annotation example} from our Sensory Ad dataset.}
    \label{fig:partial_sensations}
\end{figure}
\subsection{Sensory Ad Dataset}

\textbf{Sensation Hierarchy (Taxonomy).} 
Some advertisements are designed to evoke sensations that help audiences imagine a specific situation and the need for a product more vividly, an important factor in ad effectiveness \cite{krishna2016power}. 
In this work, we formalize the notion of sensation using a hierarchical taxonomy (partly shown in Fig.~\ref{fig:partial_sensations}; complete hierarchy in supplement). At the top level (left in the figure), our taxonomy corresponds to the five primary senses. Each is further subdivided into fine-grained categories, e.g., ``Touch'' is refined into ``Texture'', ``Temperature,'' ``Moisture and Dryness,'' ``Pain and Relief,'' and ``Pressure'', resulting in 96 total sensation labels. By definition, if an image evokes a child sensation (e.g., ``Temperature''), it also evokes its parent (e.g., ``Touch''). 
Organizing sensations in this hierarchical manner allows us to capture these dependencies and provides a structured representation for modeling and evaluating sensory understanding.
We introduce a dataset of both real and generated ads annotated with (i) up to three groups (leaves and ancestors) of sensations evoked by each image, (ii) a score reflecting the strength of evocation, and (iii) the visual elements that contribute to the sensation. 

\textbf{Data Collection.} We first selected 670 images from the PittAd dataset~\cite{PittAd}, including 250 public service advertisements (designed to raise awareness about societal issues or influence behavior) and 420 commercial advertisements (promoting products or services) to ensure a diverse range of sensory content. We have included the data statistics including the topics diversity, sensations diversity, and human-human agreement in supp.
Annotation was carried out by trained crowd workers on Prolific and using forms created on Qualtrics. 

Before contributing, each annotator was approved/filtered based on completing a practice form after reading detailed instructions, definitions of sensations, and illustrative examples. 
The annotation task followed a structured protocol: annotators first chose the most prominent sensation among the five top-level categories (with the option of selecting ``None'' if no sensation was evoked). Based on their choice, they were presented with progressively narrower subcategories until reaching a leaf-level sensation. For each selected sensation, annotators provided a strength score and listed the visual elements (e.g., colors, objects, textures) that contributed to it, using free-form text (which can be used in future work). This process was repeated up to three times per image unless ``None'' was chosen. 

The total number of sensations (at any hierarchy level) is 96, so we obtain 96 labels per image (most of which equal to 0 as the sensation is not chosen). 

For about 10\% of samples, we collected annotations from multiple workers, in order to compute inter-annotator (human-human) agreement. 
Kappa agreement \cite{kappa} was high at 0.83 with 95\% confidence interval of [0.831, 0.838].
The full annotation and testing forms are provided in the supplement, and the dataset will be released upon acceptance.

We primarily evaluate the quality of our SenseScore metric on real ad images. To test its generalizability to AI-generated advertisements, we also annotated a subset of generated ads. 
We used the action-reason statements (from \cite{PittAd}) and three annotated sensations (described above) as inputs to five 
T2I models: Flux~\cite{FLUX}, AuraFlow~\cite{AuraFlow}, PixArt~\cite{PixArt}, Stable Diffusion 3~\cite{SD3}, and Qwen-Image~\cite{QwenImage}. 
From 750 images generated by each model, we randomly selected 15 and annotated them using the same procedure as for real ads.
Table~\ref{tab:full-table-agreement} shows consistent trends, confirming the effectiveness of SenseScore on both real and synthetic ads. 
 
\subsection{Sensation Classification Tasks (SenseClass)}
Interpreting sensory ads and evaluating their effectiveness hinges on understanding which sensations an image evokes and with what intensity. Moreover, certain sensations (e.g., pain) can be inappropriate for some audiences (e.g., children), making it important for automated systems to recognize the sensations conveyed by visual content.
To formalize this, we introduce the Sensation Classification (SenseClass) task. 
We consider two 
formulations, hierarchical and single-level.

\textbf{Hierarchical Classification.} In this setting, we capture the hierarchical dependencies between sensations. 
Data annotation proceeds level by level: starting from the top-level categories, annotators choose the sensation best evoked by the image, then move to its subcategories, and so on until reaching a fine-grained leaf. The hierarchical classification task mirrors this process. Given an image, the goal is to predict the complete sensation path(s) from the root to the leaf node (e.g., Touch $\rightarrow$ Temperature $\rightarrow$ Freezing Cold). 
A model is recursively prompted to predict up to three sensations, advancing down the hierarchy by selecting among the children of each previously chosen node. 
To provide sufficient context, the definition of each potential sensation was included in the prompt.

\textbf{Single-Level Classification.}
This task flattens the taxonomy and treats every sensation, regardless of its level, as a potential label. The goal is to predict the complete set of sensations that an image evokes. 
A critical constraint in this task is maintaining hierarchical consistency. By definition, if an image evokes a specific sensation, it must also evoke its parent sensation (e.g., if ``Temperature'' is evoked, ``Touch'' is evoked as well). To evaluate a models' understanding of these relationships, we define an additional metric: Parent Recall ($R_{parent}$), which measures the fraction of predicted non-root sensations for which the direct parent sensation was also predicted. It is formally defined as:
\begin{equation}
    R_{parent} = \frac{|\{s \in S_{pred} \mid s \text{ is not a root node and } parent(s) \in S_{pred}\}|}{|\{s \in S_{pred} \mid s \text{ is not a root node}\}|}
\end{equation}
where $S_{pred}$ is the set of sensations predicted by the model. A high $R_{parent}$ score indicates that the model understands the hierarchical dependencies of sensations.

\subsection{Sensation Evocation Scoring (SenseScore)}

\begin{figure}[!tp]
    \centering
    \includegraphics[width=0.9\linewidth]{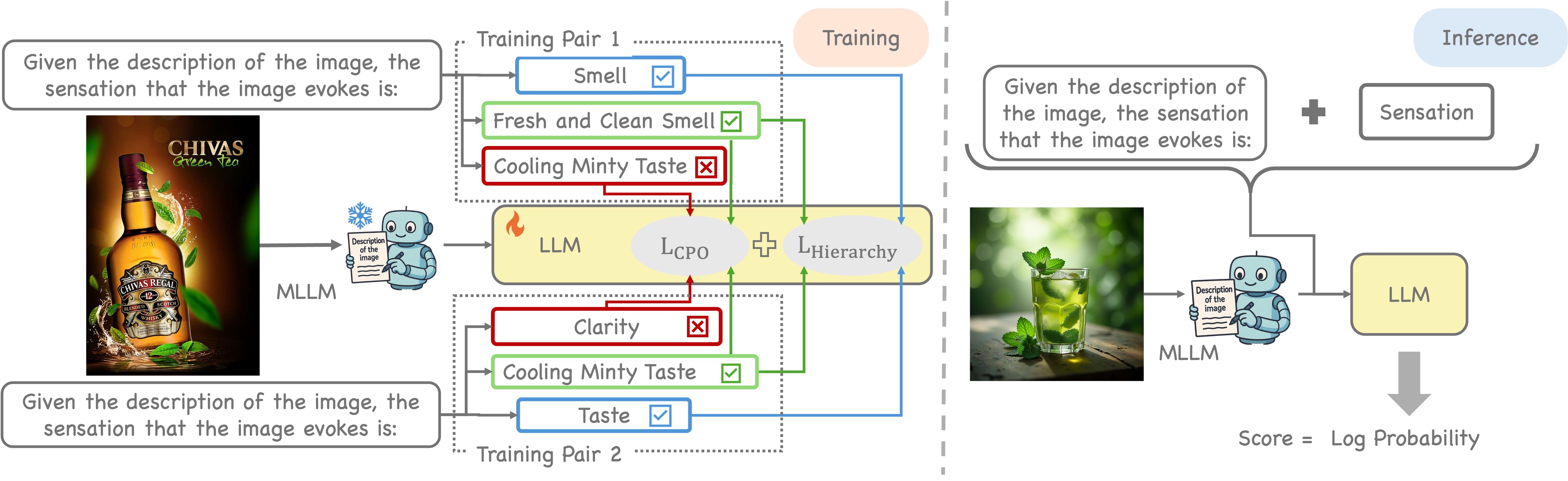}
    \caption{\textbf{SenseScore evaluation method.} Left: \colorbox{light_Apricot}{training} of LLM with two different sets of sensations for one image. 
    \textcolor{winner_green}{Green border} shows the winner sensation, \textcolor{RoyalBlue}{blue border} represents the parent of the winner (used in hierarchy loss), and \textcolor{loser_red}{red border} denotes the loser in the pair. 
    Each pair is derived from a triplet of annotations, where A is preferred over B, and B over C.  
    Right: score computation in \colorbox{inference_blue}{inference} with the fine-tuned LLM. }
    \label{fig:SenseScore}
\end{figure}

Sensation evocation can 
make ads more persuasive by enabling viewers to vividly picture the intended context \cite{elder2022review}. To quantitatively assess this effect, it is not sufficient to simply identify which sensations are present; it is also crucial to evaluate their intensity. To address this, we introduce \textbf{SenseScore}, 
which uses two stages: (i) \textbf{Image Description Generation}, where an MLLM (e.g., InternVL) generates a textual description of the image, and (ii) \textbf{Sensation Intensity Scoring}, where an LLM is prompted with the template \textit{``Given the description of the image, the sensation that the image evokes is: ''} and the average log-probability of producing the target sensation is reported as the sensation intensity score. 

Initial experiments using zero-shot LLMs show low agreement with human annotations, both in retrieving correct sensations and estimating their intensity. To address this, we fine-tune the models using a subset of our annotated dataset. In our task some sensations are evoked more than others; for example, in Fig. \ref{fig:partial_sensations}, 
both \textit{Taste} and \textit{Smell} are evoked by the image, 
but \textit{Smell} is evoked more 
strongly. A standard supervised fine-tuning approach treats both sensations as equally correct. In contrast, if sensations are paired and a model is asked to choose, \textit{Smell} should be preferred over \textit{Taste}, and \textit{Taste} should be preferred over \textit{Sight}. 
To capture such relative preferences while respecting the hierarchical structure of sensations, inspired by ~\cite{CPO}, we train SenseScore using a hierarchy-aware contrastive preference objective that encourages the model to rank sensations according to their relative strength while maintaining consistency between parent and child sensations.
\begin{equation}
\begin{split}
L_{\text{CPO+Hierarchy}}
&= -\log \sigma\!\left(\beta\big[\log \pi_\theta(y^{+}\mid x)-\log \pi_\theta(y^{-}\mid x)\big]\right) \\
&\quad + \operatorname{ReLU}\!\left(\log \pi_\theta(y^{+}\mid x)-\log \pi_\theta(y^{\text{parent}}\mid x)\right).
\end{split}
\end{equation}
where $x$ is input (prompt), $y^{+}$ is preferred output, $y^{-}$ is rejected output, $y^{parent}$ is parent of chosen output, $\pi_\theta(y \mid x)$ is the model’s conditional probability of $y$ given $x$, $\beta$ is temperature scaling, and $\sigma(\cdot)$ is the logistic sigmoid function. $L_{CPO+Hierarchy}$ encourages the model to choose $y^+$ over $y^{-}$ and prevent the probability of $y_{parent}$ from being lower than $y^+$.
We illustrate in Fig.~\ref{fig:SenseScore}.

\subsection{Sensory Ad Generation (SenseGen)}
Generative models' capability in generating images that evoke specific sensations, remains heavily unexplored.
To address this gap, we introduce the \textbf{Sensory Ad Generation} task where the input consists of an advertisement message (action-reason statement~\cite{PittAd}) and a target sensation. The objective is to generate an image that effectively conveys the message while also evoking the specified sensation. Examples outputs from prior models are shown in Fig.~\ref{fig:intro}.

\begin{figure}[t]
    \centering
    \includegraphics[width=1\linewidth]{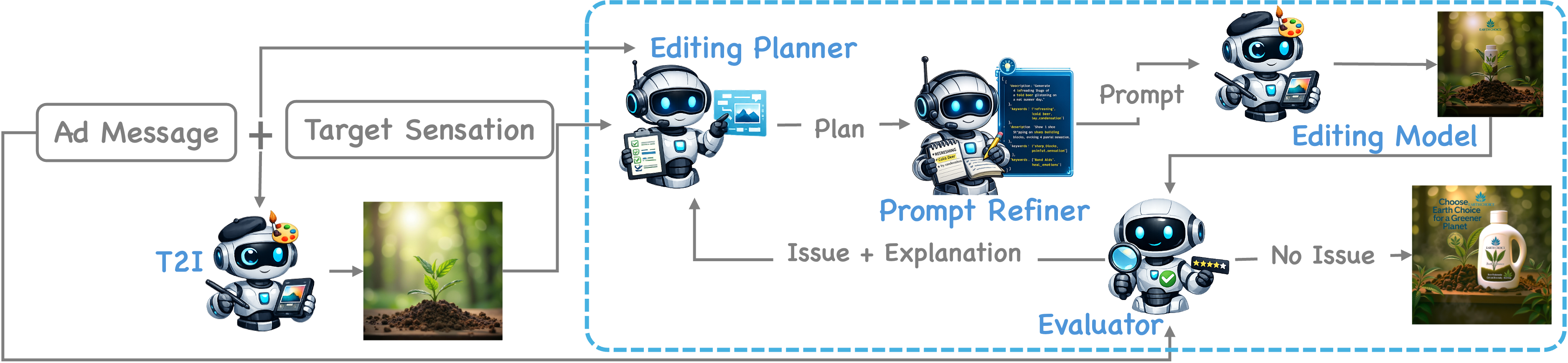}
    \caption{\textbf{SAGA framework.} Generation and editing process in multi-agent framework. \textcolor{CornflowerBlue}{Blue box} represents the conversation loop among the agents, which repeats until reaching a maximum number of messages or `No Issue' from the Evaluator Agent.}
    \label{fig:SAGA}
\end{figure}

\noindent \textbf{Sensory Ad Generation with Agent (SAGA).}
Generating sensory advertisements requires simultaneously conveying the advertisement message while evoking a target sensation through appropriate visual cues. In practice, achieving this balance often requires iterative refinement: generated images may exaggerate sensory elements or fail to clearly communicate the intended message. To address this challenge, we adopt SAGA, an agent-based iterative editing strategy that transforms generation into a structured refinement loop. Instead of relying on a single prompt-to-image step, specialized agents analyze the generated image, diagnose its shortcomings, and propose targeted edits to improve message alignment and sensation evocation, steps that would otherwise require manual iteration by human designers or specialized models.

SAGA, as illustrated in Fig.~\ref{fig:SAGA}, consists of three agents: an \emph{Editing Planner}, a \emph{Prompt Refiner}, and an \emph{Evaluator}, along with T2I and image editing models. The process begins by using a T2I model to produce an initial image conditioned on the message and the desired sensation. This image then serves as the starting point for an iterative refinement loop.
All prompts used for the agents, and an example of full conversation loop are included in the supplement.

\textbf{Editing Planner Agent} starts with the generated image. Given the image, target sensation, and advertisement message, the agent produces a structured list of edit actions (e.g., adding or removing visual elements, modifying attributes such as color, texture, or style) aimed at improving both conveying message and evoking the target sensation.
After each iteration, the editing planner receives feedback from the Evaluator Agent on visual consistency, alignment with ads message, and how effectively the target sensation is evoked. Using this feedback, along with the edited image and original inputs, the planner generates updated actions to address the identified issue and progressively improve the image.

\textbf{Prompt Refiner Agent} converts the Planner’s structured action list (provided in JSON format) into a coherent editing prompt for the image editing model. It merges redundant actions, resolves inconsistencies, and ensures that the final prompt clearly reflects the intended modifications. The refined prompt is then used as input to the editing model to produce the next image.

\textbf{Evaluator Agent} assesses each generated image along three dimensions: (1) visual element consistency, (2) image–message alignment, and (3) sensation evocation. 
First, the agent is given detailed instructions how to check for visual inconsistencies (incoherent or conflicting visuals such
as artifacts, glitches, or contradictory elements); if such issues are detected, it flags visual consistency as the primary problem and provides an explanation of what the inconsistency is. If the image is visually consistent, the evaluator agent next examines whether the advertisement message is clearly conveyed, prioritizing alignment since preliminary results indicate that T2I models may exaggerate sensory cues ignoring the advertisement message. If the critic deems the message is not clearly conveyed (the product is not prominent, or the image does not reinforce the message), the agent 
specifies which parts of message are missing or unclear. Finally, if 
the target sensation is deemed weak or not
effectively evoked through visual cues, colors, lighting, objects, or atmosphere, the agent 
explains the deficiency. If all three criteria are satisfied, the Evaluator returns ``No Issue'' and the editing process terminates.

\section{Results}
We begin by benchmarking LLMs and MLLMs on our sensation classification tasks to assess their understanding of sensory concepts. We then validate our proposed SenseScore metric, comparing against baseline metrics. Finally, we evaluate the performance of leading T2I models on the SenseGen task and compare it with our proposed framework (SAGA). Implementation details are in supp.


\begin{table}[t]
\caption{Results on classification. ``-'' denotes model did not follow instructions. }
\centering
\scriptsize
\begin{tabular}{c|c|c|c|c|c|c|c}
\multirow{2}{*}{\textbf{Model}} &
\multicolumn{3}{c|}{\textbf{Hierarchical Classification}} &
\multicolumn{4}{c}{\textbf{Single-Level Classification}} \\\cmidrule{2-8}

&\textbf{$P$} &\textbf{$R$} &\textbf{$F1$} &
\textbf{$P$} &\textbf{$R$} &\textbf{$F1$} &
\textbf{$R_{parent}$}\\\midrule

\multicolumn{8}{l}{\textbf{MLLMs}}\\\midrule

QwenVL &
0.17 & 0.62 & 0.27 &
\textbf{0.33} & 0.18 & 0.23 &
0.45\\

InternVL &
0.13 & 0.60 & 0.21 &
0.18 & 0.44 & 0.26 &
0.41\\

LLAVA-Next &
0.10 & 0.60 & 0.17 &
- & - & - &
- \\

GEMMA &
0.17 & \textbf{0.66} & 0.27 &
0.11 & 0.39 & 0.17 &
0.49\\

\hline

\multicolumn{8}{l}{\textbf{LLMs}}\\\midrule

QwenLM + $D_{QwenVL}$ &
0.18 & 0.45 & 0.26 &
0.18 & 0.42 & 0.25 &
0.24\\

QwenLM + $D_{InternVL}$ &
0.18 & 0.44 & 0.26 &
0.18 & 0.42 & 0.25 &
0.22\\

QwenLM + $D_{GEMMA}$ &
0.18 & 0.45 & 0.26 &
0.19 & 0.44 & 0.27 &
0.24\\

GEMMA + $D_{QwenVL}$ &
0.16 & 0.54 & 0.25 &
0.13 & \textbf{0.54} & 0.21 &
0.65\\

GEMMA + $D_{InternVL}$ &
0.15 & 0.54 & 0.23 &
0.13 & \textbf{0.54} & 0.21 &
0.64\\

GEMMA + $D_{GEMMA}$ &
0.15 & 0.55 & 0.24 &
0.14 & \textbf{0.54} & 0.22 &
\textbf{0.68}\\

LLAMA3 + $D_{QwenVL}$ &
0.19 & 0.43 & 0.26 &
0.15 & 0.47 & 0.23 &
0.48\\

LLAMA3 + $D_{InternVL}$ &
\textbf{0.21} & 0.43 & \textbf{0.28} &
0.13 & 0.47 & 0.20 &
0.45\\

LLAMA3 + $D_{GEMMA}$ &
0.20 & 0.43 & 0.27 &
0.13 & 0.48 & 0.20 &
0.46\\

\end{tabular}
\label{tab:retrieval}
\end{table}

\subsection{Sensation Classification Tasks}
Our evaluation follows distinct protocols based on the model's input modality. For MLLMs, 
the ad image was provided as direct visual input. The model was then tasked with classifying the corresponding sensations based on a task-specific prompt (see supp). 
To assess the performance of text-only LLMs,
we employed a two-stage pipeline. First, we utilized different MLLM (InternVL, QwenVL, and Gemma) to generate a description for the image ($D_{MLLM}$). These generated descriptions were utilized as input context for the LLMs to perform sensation classification. 
This approach allows us to isolate and evaluate the language-based reasoning capabilities of LLMs for this specific task. We report Recall (R), Precision (P), and F1-score (F1). For Single-Level Classification, we also report the Parent Recall ($R_{parent}$) to assess understanding the hierarchical relations.

\textbf{Hierarchical Classification.} 
Table \ref{tab:retrieval} reveals a consistent trend across all models: significantly higher recall than precision. This imbalance indicates that while models are proficient at identifying potentially relevant sensations, they struggle to reject incorrect ones.
The best F1 numbers are found among MLLMs, and MLLMs often outperform their LLM counterparts on this task (e.g., QwenVL / QwenLM, GEMMA / GEMMA). 
This suggests that direct visual input provides crucial cues that may be lost or distorted in text-only descriptions. 

\textbf{Single-Level Classification.} 
The results in Table \ref{tab:retrieval} show that while MLLMs achieve higher precision and F1-scores, some LLMs (GEMMA) have a stronger performance on Parent Recall ($R_{parent}$). This suggests that 
LLMs, operating on textual descriptions and definitions, develop a better understanding of the abstract, semantic relationships between sensations in the hierarchy.

\begin{table}[!tp]
    \caption{Kappa agreement between human annotators and evaluation metrics 
    on 100 real images ($\sim$10,000=100x96 image-sensation pairs) and 50 generated ads ($\sim$5,000 image-sensation pairs). CI = confidence interval.} 
    \centering
    \scriptsize
    \setlength{\tabcolsep}{1pt}
    \begin{tabular}{c||c|c|c|c|c|c|c||c}
        \multirow{2}{*}{Metrics} & \multicolumn{7}{c||}{Real Ads} &   \multirow{2}{*}{Gen. Ads} \\
        & Touch & Smell & Sound & Taste & Sight & All & 95\% CI & \\
       \hline 
       \multicolumn{9}{l}{Baselines}\\
       \hline
       VQA-score & 0.58 & 0.60 & 0.42 no& 0.65 & 0.58 & 0.57 & [0.56, 0.57] & 0.52\\
       Image-Reward & 0.49 & 0.50 & 0.38 & 0.34 & 0.45 & 0.46 & [0.45, 0.46] & 0.40 \\
       CLIP-score & 0.48 & 0.47 & 0.36 & 0.41 & 0.30 & 0.44 & [0.43, 0.44] & 0.45\\
       Pick-score & 0.38 & 0.45 & 0.12 & 0.36 & 0.30 & 0.36 & [0.35, 0.36] & 0.41\\ 
       \hline       
       \multicolumn{9}{l}{LLM/MLLM as a judge}\\
       \hline
       InternVL& 0.54 & 0.48 & 0.43 & 0.54 & 0.49 & 0.50 & [0.50, 0.51] & 0.48 \\
       QwenVL & 0.55 & 0.48 & 0.43 & 0.54 & 0.50 & 0.50 & [0.50, 0.51] &  0.43 \\
       LLAMA3 + $D_{InternVL}$ & 0.37 & 0.38 & 0.39 & 0.48 & 0.52 & 0.48 & [0.47, 0.48] & 0.47 \\
       QwenLM + $D_{InternVL}$ & 0.30 & 0.30 & 0.28 & 0.54 & 0.52 & 0.45 & [0.45, 0.46] & 0.48 \\
       \hline 
       \multicolumn{9}{l}{Zero-shot SenseScore}\\
       \hline
       LLAMA3 + $D_{InternVL}$ & -0.09 & 0.08 & -0.22 & -0.01 & -0.01 & -0.03 & [-0.03, -0.02] & -0.01\\
       QwenLM + $D_{InternVL}$ & -0.15 & 0.04 & -0.22 & 0.03 & 0.003 & -0.06 & [-0.06, -0.05] & -0.04\\ 
       \hline 
       \multicolumn{9}{l}{SenseScore}\\
       \hline
       SenseScore (LLAMA3 + $D_{InternVL}$) & \textbf{0.79} & \textbf{0.82} & \textbf{0.77} & \textbf{0.84} & \textbf{0.85} & \textbf{0.80} & [0.80, 0.81] & \textbf{0.68}\\
       SenseScore (LLAMA3 + $D_{QwenVL}$) & 0.76 & 0.77 & 0.70 & 0.79 & 0.73 & 0.75 & [0.75, 0.76] & 0.67\\
       SenseScore (QwenLM + $D_{InternVL}$) & 0.64 & 0.69 & 0.57 & 0.73 & 0.64 & 0.66 & [0.65, 0.66] & 0.56\\
       SenseScore (QwenLM + $D_{QwenVL}$) & 0.62 & 0.66 & 0.50 & 0.67 & 0.58 & 0.61 & [0.61, 0.62] & 0.55\\
       
    \end{tabular}
    \label{tab:full-table-agreement}
\end{table}

\subsection{Sensation Evocation Scoring}
To evaluate the accuracy of our metric, we use about 10,000 human annotations (image-sensation pairs over 100 images separate from those used for training SenseScore, and 96 sensation labels) plus about 5,000 annotations on generated images. The intensity for each sensation is set to the score chosen by the annotator (or 0 if not chosen). 
We report Kappa ($\kappa$), where we use the sensation with higher score as the chosen one. We also show Pearson ($r$) 
in supp.

\textbf{SenseScore compared to baselines.} We benchmark SenseScore against baseline metrics, including VQA-score \cite{lin2024evaluating}, ImageReward \cite{ImageReward}, CLIP-score \cite{CLIP-score}, and Pick-score \cite{PickScore}. To demonstrate the necessity of our proposed fine-tuning procedure, we further compare SenseScore against the zero-shot performance of the SenseScore inference pipeline using LLAMA3-instruct ($LLAMA3$) and QwenLM ($QwenLM$) with image descriptions generated by InternVL ($D_{InternVL}$) and QwenVL ($D_{QwenVL}$). As observed in Table \ref{tab:full-table-agreement}, among baseline metrics, VQA-score achieves the highest human-metric agreement with moderate performance ($\kappa$ = 0.57
on real ads and $\kappa$ = 0.52 
on generated images). In contrast, fine-tuned SenseScore reaches near-perfect agreement with human ($\kappa$ = 0.85) 
on real ads and substantial performance ($\kappa$ = 0.68) 
on generated ads, representing a 49\% and 31\% improvement on Kappa agreement, respectively.
Notably, zero-shot variants of SenseScore---LLAMA3 (zero-shot) and QwenLM (zero-shot)---exhibit 
complete misalignment with human judgments, emphasizing that our fine-tuning procedure is 
essential for alignment with human perception and the better performance of our metric. 
This result also shows the superior performance of SenseScore is not the result of information leakage from description generation (since the zero-shot methods also use descriptions but perform poorly).
We further use MLLMs, InternVL and QwenVL, as a judge for sensation evocation; these models both show inferior performance to our method. 

\begin{figure}[!t]
    \centering
    \includegraphics[width=1\linewidth]{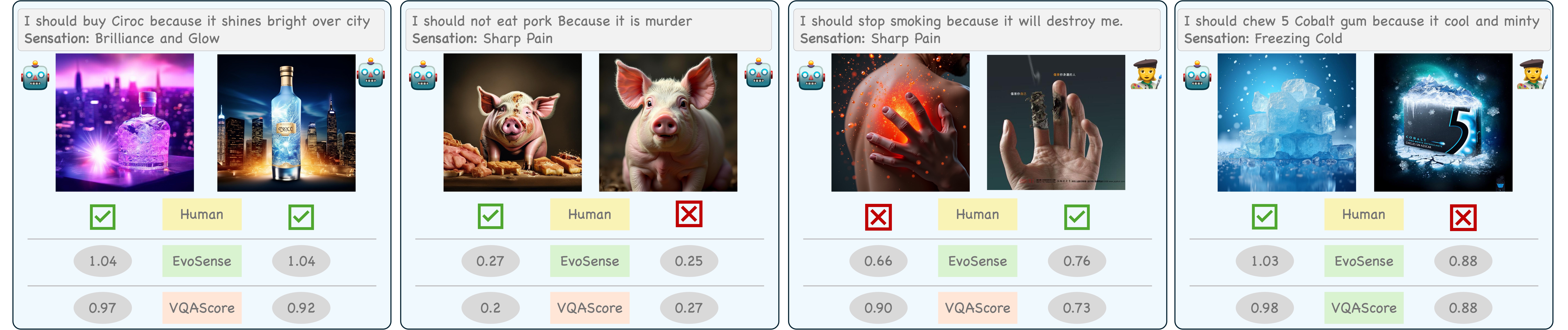}
    \caption{\textbf{Examples on human agreement with SenseScore and VQA-score}  
    on intensity of sensations. The Human row shows the chosen 
    (\textcolor{Green}{$\checkmark$}) 
    image(s) (including ties) and rejected (\textcolor{red}{$\times$}) image. \colorbox{light_Apricot}{Red background} indicates the model-chosen (higher-scoring) option is misaligned with human choice, and \colorbox{first_level_green}{green background} shows it is aligned. }
    \label{fig:metrics_example}
\end{figure}

The examples in Fig.~\ref{fig:metrics_example} show higher agreement of SenseScore with human annotation compared to VQA-score (the best baseline).

\textbf{Ablation on SenseScore.} We 
analyze the impact of the core components of SenseScore: the base LLM and the MLLM used for description generation. The results in Table \ref{tab:full-table-agreement} show that while both fine-tuned LLMs significantly outperform all baseline metrics, LLAMA3-instruct holds a slight edge over QwenLM in human agreement. Further, the results demonstrate the robustness of our method to the source of image descriptions. When the descriptions are generated by QwenVL ($D_{QwenVL}$) instead of InternVL ($D_{InternVL}$), the change in agreement scores for the fine-tuned models is minimal. 

\subsection{Sensory Ad Generation}

\begin{table}[t]
    \caption{Evaluating generated sensory ads.
    ``-'' indicates metric does not apply.
    }
    \centering
    \scriptsize
    \setlength{\tabcolsep}{4pt}
    \begin{tabular}{c||c|c|c|c}
         \multirow{2}{*}{T2I model} & \multicolumn{4}{c}{Sensory Ad} \\
         \cline{2-5}
         & Input &SenseScore & AIM & $P_{comp}$ \\
         \hline
         \hline
         Flux & AR & - & 0.43 & 0.54 \\
         Flux & AR + Sensation & 0.97 &  0.39 & 0.56 \\
         SD3 & AR + Sensation & 0.96  & 0.42 & 0.59 \\
         AuraFlow & AR + Sensation & 0.96 &  0.39 & 0.57 \\
         PixArt & AR + Sensation & 0.96 & 0.39 & 0.61 \\
        Qwen-Image & AR + Sensation & 0.98 & 0.43 & 0.57 \\
         DALLE-3 & AR + Sensation & 0.98 &  0.45 & 0.61 \\
        \hline
        SAGA (FLUX) & AR & - & 0.46 & 0.61 \\
        SAGA (FLUX)  & AR + Sensation & \textbf{0.99} & \textbf{0.49} & \textbf{0.62} \\
         
    \end{tabular}
    \label{tab:AdEval}
\end{table}

First, we benchmark different T2I models including Flux \cite{FLUX}, Stable Diffusion 3 (SD3) \cite{SD3}, AuraFlow \cite{AuraFlow}, PixArt \cite{PixArt}, and Qwen-Image \cite{QwenImage}, on the SenseGen task evaluating their abilities in generating images that convey specific ad messages and evoke the given sensation to make the images more persuasive. 
We test performance when just the ad message is fed as input (AR) as opposed to message and sensation (AR + Sensation).
In supp, we also 
benchmark generating images that evoke sensations without an ad message.

\begin{figure}[!t]
    \centering
    \includegraphics[width=0.9\linewidth]{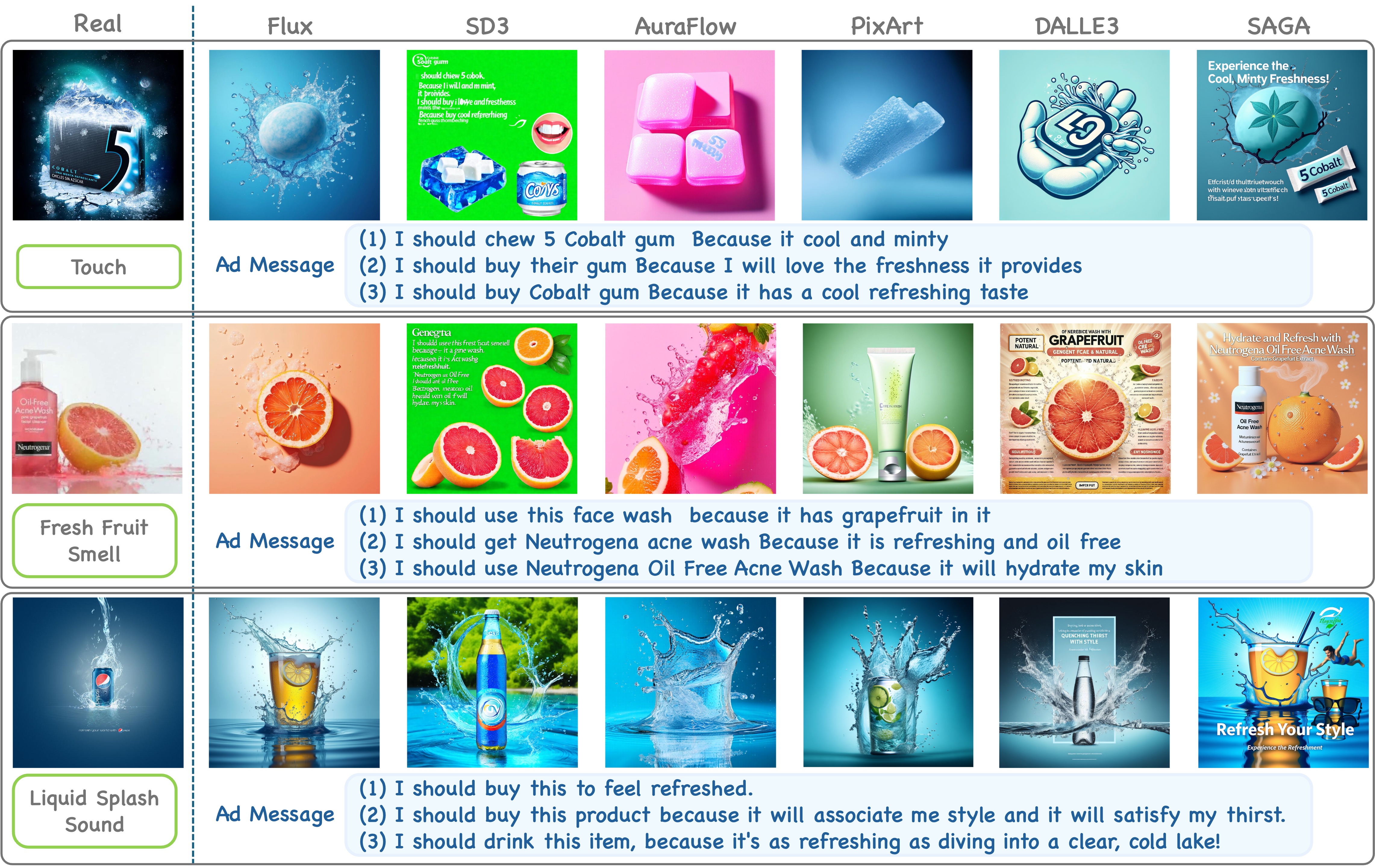}
    \caption{\textbf{Sensory Ad examples.} Three examples of real ads and ads generated by Flux \cite{FLUX}, SD3 \cite{SD3}, AuraFlow\cite{AuraFlow}, PixArt \cite{PixArt}, DALLE3 \cite{DALLE3} and SAGA (ours) given the action-reason message 
    and sensation annotation for the real advertisement. 
    \textcolor{YellowGreen}{Green border} represents the sensation used in the prompt of T2I models.}
    \label{fig:genvsreal}
\end{figure}


\begin{figure}[t]
    \centering

    \begin{subfigure}[t]{0.45\linewidth}
        \centering
        \includegraphics[width=1\linewidth]{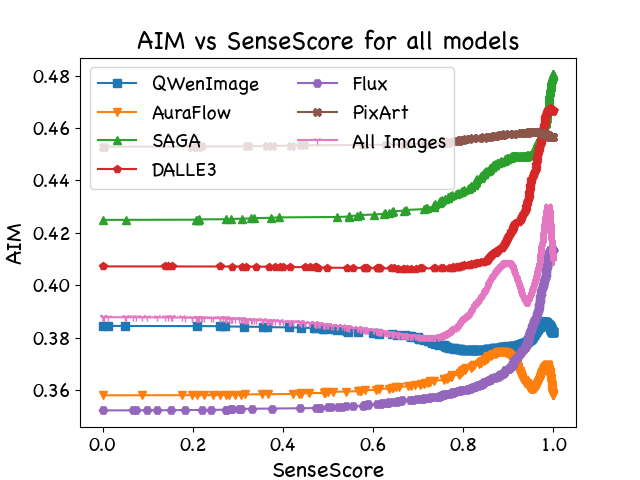}
            \caption{Relation between Sensation Evocation and Text-Image Alignment in generated ads.}
        \label{fig:exaggeration_plot}
    \end{subfigure}
    \hfill
    \begin{subfigure}[t]{0.45\linewidth}
        \centering
    \includegraphics[width=1\linewidth]{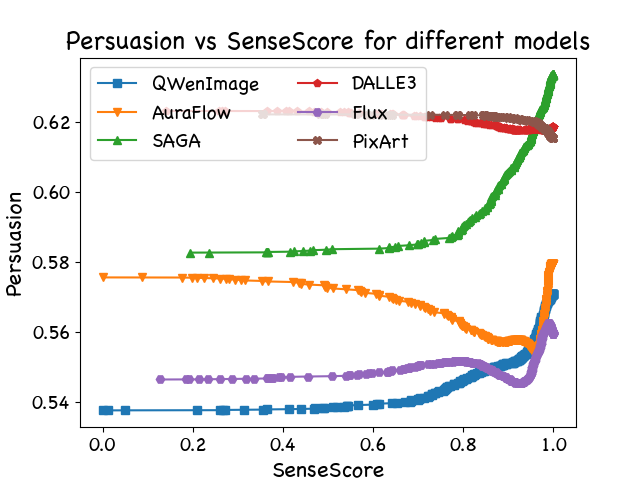}
    \caption{Relation between sensation and persuasion.}
    \label{fig:sens_persuasion}
    \end{subfigure}
    \caption{Analysis of the relation between sensation intensity, text-image alignment, and persuasion in generated and real advertisements.}
    \label{fig:sensation_analysis}
\end{figure}

\textbf{Sensory Ad Generation Performance.} 
Table \ref{tab:AdEval} highlights the higher performance of our proposed method, and \textbf{introduces its impact on ad effectiveness}.  
In addition to our SenseScore, we report two metrics from prior work \cite{CAP}: AIM measures alignment of the generated image with the intended ad message, and $P_{comp}$ measures persuasiveness through multiple components (questions); both were shown to agree well with human judges. 
Table \ref{tab:AdEval} shows that among existing T2I models, Qwen-Image and DALLE-3 achieve the highest sensation intensity.
Our proposed SAGA performs on par, slightly exceeding the intensity. Importantly, it also \textbf{outperforms other methods in terms of alignment with the intended message (AIM) and persuasion ($P_{comp}$).} 


Fig.~\ref{fig:genvsreal} shows qualitative comparisons between SAGA and baselines. \textbf{SAGA effectively conveys both the implicit ad message and the intended sensory experience by integrating appropriate visual cues}. In contrast, baselines often produce overly literal interpretations of sensations (e.g., DALLE3 depicts  literal ``touch'' in 1st row while our method connects touch to  freshness), over-rely on text rather than rhetoric visual cues (e.g., DALLE3 in 2nd row), or miss the core ad message entirely (e.g., AuraFlow generating a generic splash scene in 3rd row).

We note that while the goal is to evoke specific sensations, sometimes models exaggerate in evoking the sensation,
overlook the advertisement message, and only show sensation-associated objects; we discuss this shortly. This explains why higher SenseScore and higher AIM/$P_{comp}$ are not always correlated. 

\subsection{Connection of alignment, persuasion and sensation evocation}
\label{sec:per-sens}
We now analyze in more detail the relation between the alignment (AIM) of AR messages and images, the persuasion metric (both from \cite{CAP}), and sensation evocation in generated images. We plot the alignment over sensation intensity in Fig.~\ref{fig:exaggeration_plot}, and persuasion score over intensity in Fig.~\ref{fig:sens_persuasion}.
We applied a Gaussian filter to the AIM scores to smooth the visualization. 
Fig.~\ref{fig:exaggeration_plot} reveals that for several methods, alignment initially increases with sensation evocation, reaches a peak, and subsequently decreases as sensation evocation continues to increase. This suggests there is an optimal level of sensation evocation, which our method successfully finds (as shown by the superior performance on all three metrics in Table~\ref{tab:AdEval}). 
In Fig.~\ref{fig:sens_persuasion}, we observe that the persuasion score for the images generally increases with the increase in the sensation intensity. \textbf{This suggests that sensory evocation generally boosts persuasiveness.} 

\label{sec:exaggeration}
\begin{figure}
    \centering
    \includegraphics[width=0.8\linewidth]{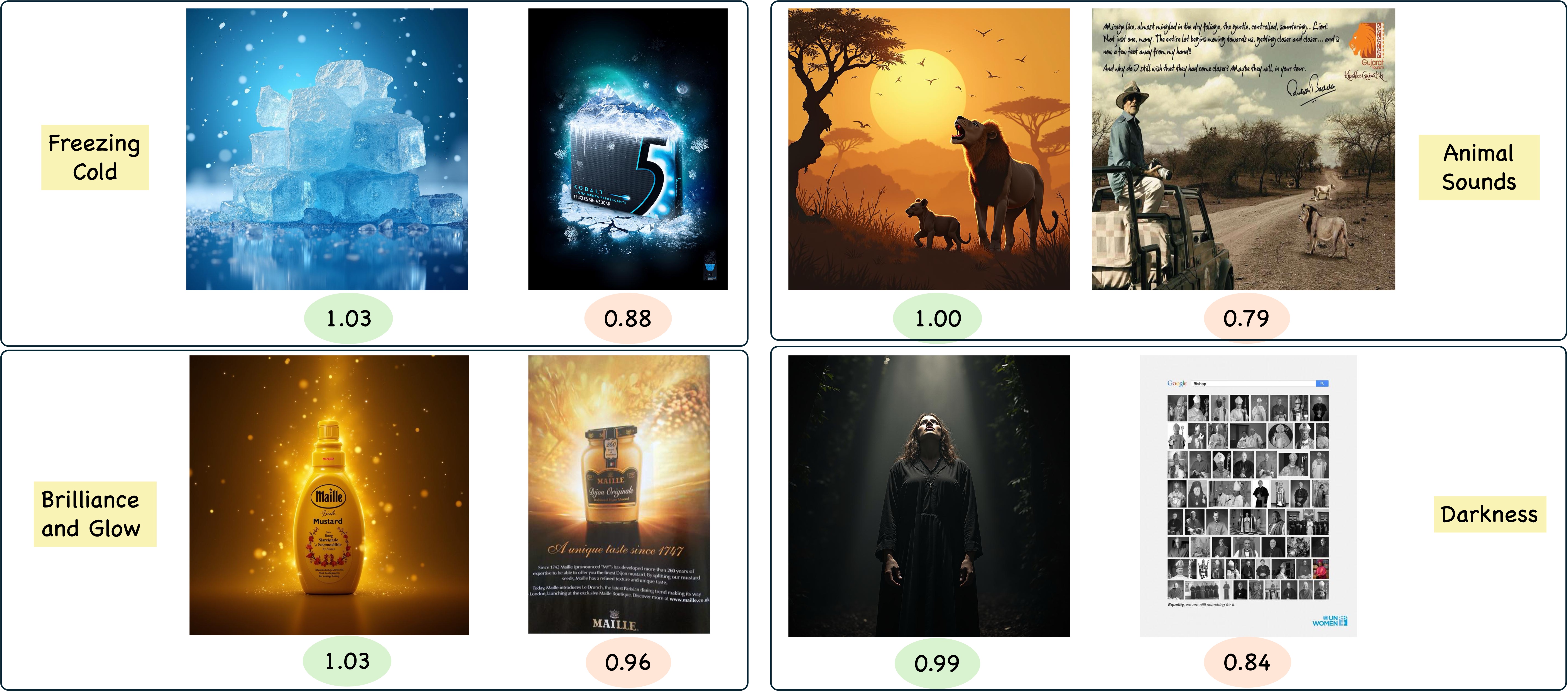}
    \caption{\textbf{Examples of exaggeration in sensation evocation.} In each group the image on the left is a generated advertisement and the one on the right is a real ad. We report the corresponding SenseScores (green is higher but not necessarily better).}
    \label{fig:exag}
\end{figure}

To understand the desirable level of sensory evocation, 
we compared the intensity of sensation in generated and real advertisements. Average sensation intensity in real ads computed by SenseScore is 0.83 (not shown in table) which is lower than intensity of sensations in the ads generated by T2I models. 
This is due to exaggeration in evoking sensations 
commonly associated with some objects, like `Freezing Cold' often represented by ice-cubes or snow.
In Fig.~\ref{fig:exag}, generated images evoke the input sensation with higher intensity than real ads; however, this exaggeration 
results in overlooking the advertisement message and failing in conveying it. For example, in Fig.~\ref{fig:exag}, the top left image is supposed to convince the audience to buy the Five gum by showing the cooling and refreshing feature of the product; however, while the image intensely evokes the sensation it fails to convey the message. Our proposed SAGA copes with over-exaggeration by receiving specific feedback and correcting its outputs.

\section{Conclusion}
We addressed the challenging, previously unexplored task of generating and understanding visual content that evokes specific human sensations, a crucial element of persuasive advertising. To facilitate research in this area, we introduced the \emph{Sensory Ad benchmark} including the Sensory Ad dataset with a detailed hierarchical taxonomy for sensations, two Sensation Classification tasks, and the new Sensory Ad Generation task. We propose \emph{SenseScore}, an evaluation metric that accurately measures the intensity of evoked sensations. By fine-tuning an LLM with a hybrid objective (CPO and hierarchical constraints), SenseScore achieves high agreement with human judgments, significantly outperforming existing baselines significantly. 
We also propose SAGA, an agentic framework that proposes and evaluates edits to an image to improve sensory evocation, message alignment, and persuasiveness. 
This work lays the foundation for developing a new generation of sensation-aware models and expanding the scope of understanding sensory content beyond advertising.

\bibliography{main}
\bibliographystyle{splncs04}

\appendix
\section{Outline}
\label{sec:appendix-outline}

This supplement provides additional discussion of the dataset, experimental setup and implementation details, and extended results. It also discusses potential ethical considerations and includes the prompts used in our implementation to ensure reproducibility (code will be released upon acceptance). An example of the survey is attached at the end of this supplement. The supplement is organized as follows:

\begin{itemize}
    \item \textbf{Dataset} (Section~\ref{sec:appendix-dataset})
    \begin{itemize}
        \item Taxonomy and data collection (Fig.~\ref{fig:sensations})
        \item Sensation and topic diversity (Fig.~\ref{fig:sensation_dist})
        \item Annotation protocol
        \item Human--human agreement ($\kappa=0.83$, 95\% CI)
    \end{itemize}

    \item \textbf{Experimental Setup \& Implementation Details} 
    (Section~\ref{sec:appendix-experimental-setup})
    \begin{itemize}
        \item Sensation classification
        \item SenseScore training
        \item Description generation (Fig.~\ref{fig:description_examples})
        \item SensoryAd generation setup
    \end{itemize}

    \item \textbf{Additional Results \& Analysis} 
    (Section~\ref{sec:supp_addtional_results}) 
    \begin{itemize}
        \item \textit{SenseScore Evaluation} (Section~\ref{sec:supp_SenseScore_eval})
        \begin{itemize}
            \item Ablation on number of fine-tuning iterations (Table~\ref{tab:fine-tuning-ab})
            \item Human agreement and baseline comparison (Table~\ref{tab:agreement})
            \item Kappa agreement and Pearson correlation gap (Fig.~\ref{fig:kappavsr})
            \item Comparison on extended number of real images (Table~\ref{tab:bigset-table-agreement})
        \end{itemize}

        \item \textit{Sensory Ad Generation} (Section~\ref{sec:supp_sensoryAds_generation})
        \begin{itemize}
            \item Sensory Ads generation results (Table~\ref{tab:supp_saga_full})
            \item Fine-tuning SD3 on SensoryAd generation
        \end{itemize}

        \item \textit{Sensory Images Beyond Ads} (Section~\ref{sec:supp_sensory_image_beyond_Ads})
        \begin{itemize}
            \item Results on sensory images beyond ads (Table~\ref{tab:SensoImgEval})
            \item Variation in generation performance across sensations (Fig.~\ref{fig:sensation_heatmap})
        \end{itemize}
    \end{itemize}

    \item \textbf{Ethical Concerns \& AI Usage} 
    (Section~\ref{sec:supp_ethic_ai_usage})
    \begin{itemize}
        \item Ethical concerns around sensory advertisements
        \item Usage of AI
    \end{itemize}

    \item \textbf{Prompts} 
    (Section~\ref{sec:supp_prompts})
 \item \textbf{Illustration of User Study} (attached at the end) 

\end{itemize}

\section{Dataset}
\label{sec:appendix-dataset}

\begin{figure}[!tp]
    \centering
    \includegraphics[width=1\linewidth]{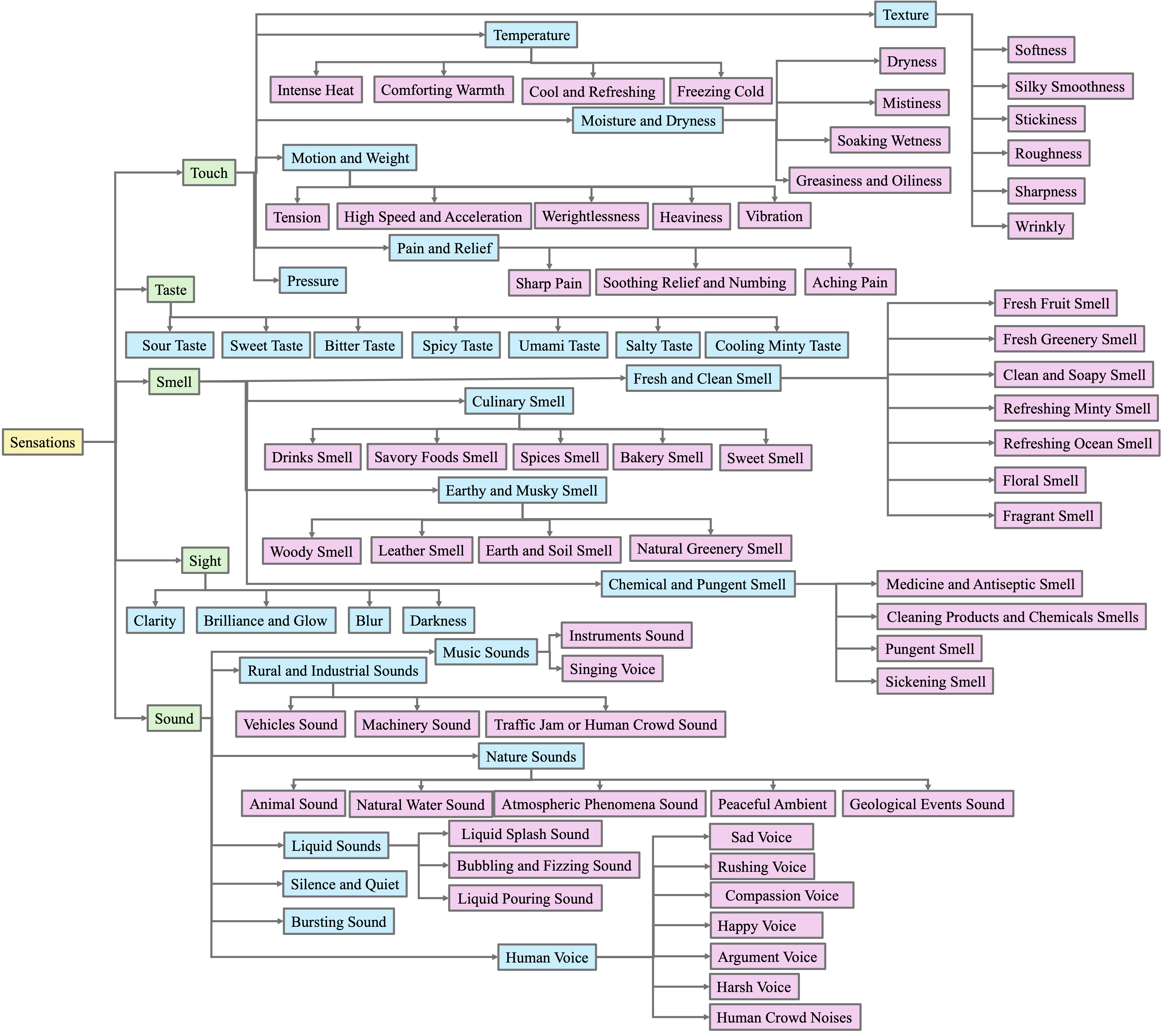}
    \caption{\textbf{Sensation Hierarchy.} First level, represents the main five human sensations, and each sensation is categorized into different set of sensations.}
    \label{fig:sensations}
\end{figure}

\textbf{Taxonomy and Data Collection.}
To collect the dataset, we first defined the taxonomy shown in Fig.~\ref{fig:sensations}. 
We then randomly sampled 670 images from the PittAd dataset~\cite{PittAd}, covering 95 sensations and more than 40 topics.

\begin{figure}[h]
    \centering

    \begin{subfigure}[t]{0.4\linewidth}
        \centering
        \includegraphics[width=\linewidth]{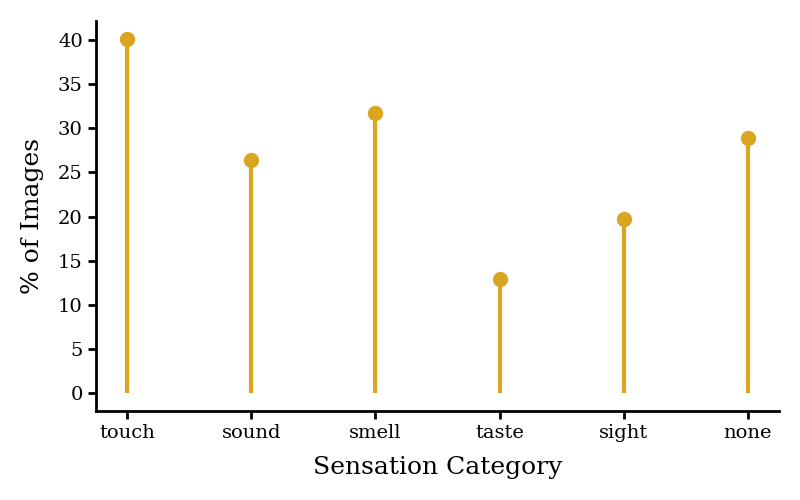}
        \caption{Diversity of images over 5 main sensations.}
        \label{fig:sensation_dist_a}
    \end{subfigure}
    \hfill
    \begin{subfigure}[t]{0.5\linewidth}
        \centering
        \includegraphics[width=\linewidth]{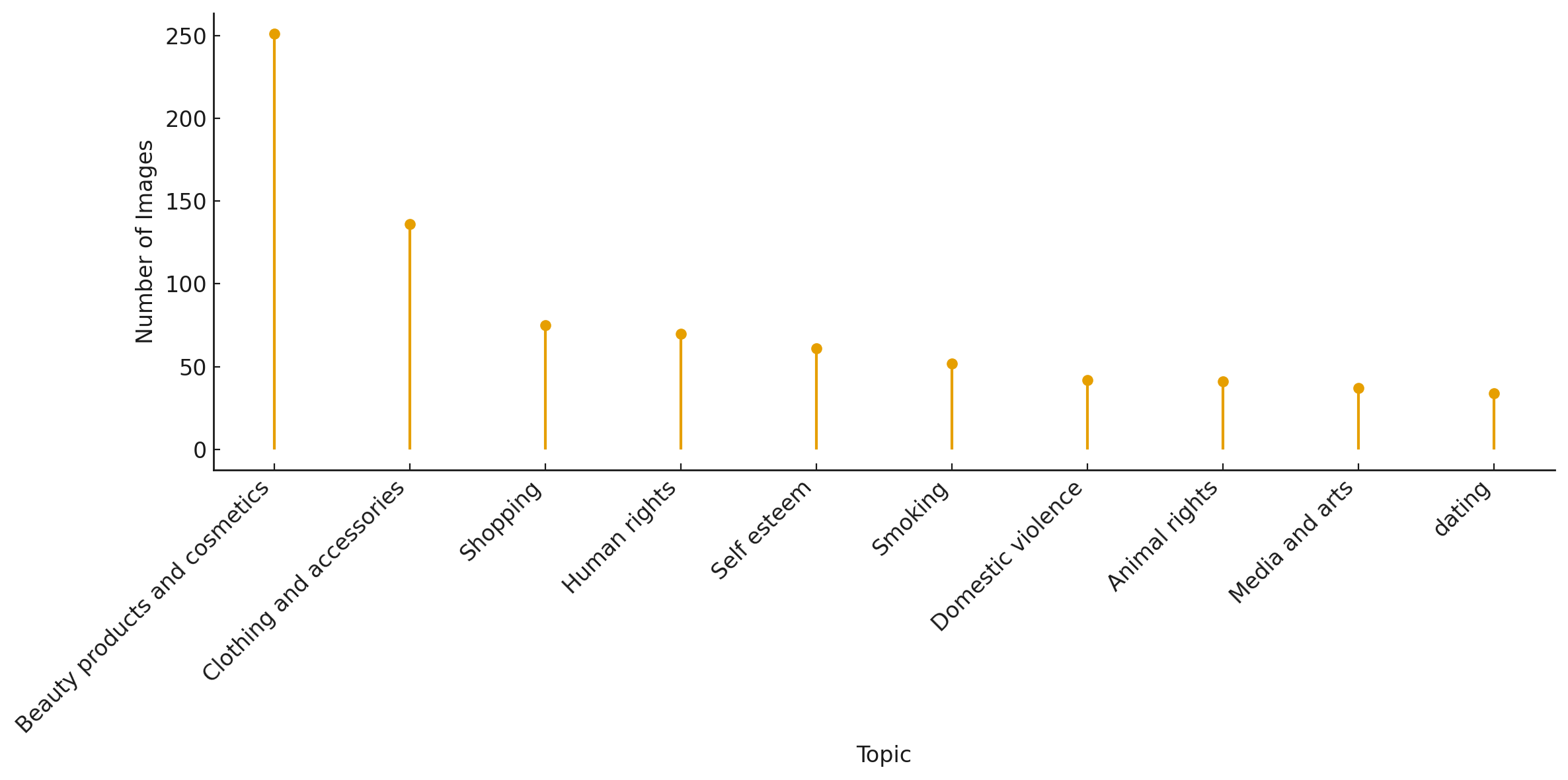}
        \caption{Diversity of images over 10 most frequent topics in the SensoryAd dataset.}
        \label{fig:sensation_dist_b}
    \end{subfigure}
    \caption{Image Distribution. Left: percentage of images per sensation category (sensation diversity). Right: distribution over the 10 most frequent topics (topic diversity).}
    
    \label{fig:sensation_dist}
\end{figure}

\textbf{Sensation and Topic Diversity.}
 Fig.~\ref{fig:sensation_dist_a} and Fig.~\ref{fig:sensation_dist_b} show the distribution over the 5 main sensations and the 10 most frequent topics, exhibiting diversity across both sensory modalities and topics. 

\textbf{Annotations.}
For data annotation, we first had a test phase study on Prolific, gave the annotators detailed instruction with examples of images evoking each sensation, and selected a group of annotators based on the quality of their responses to do the main study. 
We used Qualtrics to create dynamic forms showing different options based on annotators choice in each step. The form is uploaded as the supplementary file. 

The final annotations were collected from 12 annotators of different genders, aged 25–60, all with at least a high school diploma, an approval rate above 90\% on more than 1000 prior annotations, and located in the United States. Each image was annotated by one annotator and subsequently reviewed by a skilled evaluator for quality assurance. In cases of disagreement (which were rare), the annotator was asked to justify their choice. If the justification was insufficient, the annotation was discarded and the image was returned to the annotation pool.

\textbf{Human–Human Agreement.}
To assess reliability, we collected two independent annotations for approximately 10\% of the images and computed $\kappa$. The human–human agreement is 0.83 (95\% CI: [0.831, 0.838]), indicating strong inter-annotator consistency.

\section{Experimental Setup \& Implementation Details} 
\label{sec:appendix-experimental-setup}
In this section we explain the experimental setup. Hugging Face implementation of models are utilized, and code will be released upon the acceptance. 

\subsubsection{Sensation Classification} In sensation classification tasks, we evaluated the model on real ads images in our dataset. We benchmark MLLMs including the InternVL(InternVL3.5-8B), Gemma (gemma-3-4b-it), QwenVL (Qwen2.5-VL-7B-Instruct), and LLAVA-Next (llava-v1.6-vicuna-13b-hf) with 8-bit quantization for models with more than 4 Billion parameters. We also benchmark LLMs including Gemma, LLAMA3 (Meta-Llama-3-8B-Instruct), and QwenLM (Qwen2.5-7B-Instruct), given the descriptions generated by the same MLLMs. Similar to MLLMs 8-bit quantization was applied on models with more than 4B parameters.
\subsubsection{SenseScore Training} 
We fine-tune LLMs using LoRA \cite{LoRA} on 40000 image-sensation pairs. To train our proposed evaluation metric, we randomly selected 100 images from annotations to create our training data. In our proposed training, we pair each two sensations with different intensity (scores chosen by human annotators) as chosen and rejected. Each data point in our training, included description of the image, chosen sensation, rejected sensation, and parent of chosen sensation. This training data setting resulted in 40000 data point. We fine-tuned the LLMs - LLAMA3 (Meta-Llama-3-8B-Instruct), and QwenLM (Qwen2.5-7B-Instruct) - using LoRA \cite{LoRA} with batch-size of 1, and learning rate 5e-5. Our evaluation of SenseScore performance was on a subset of the images not selected for training.

\begin{figure}[!tp]
    \centering
    \includegraphics[width=1\linewidth]{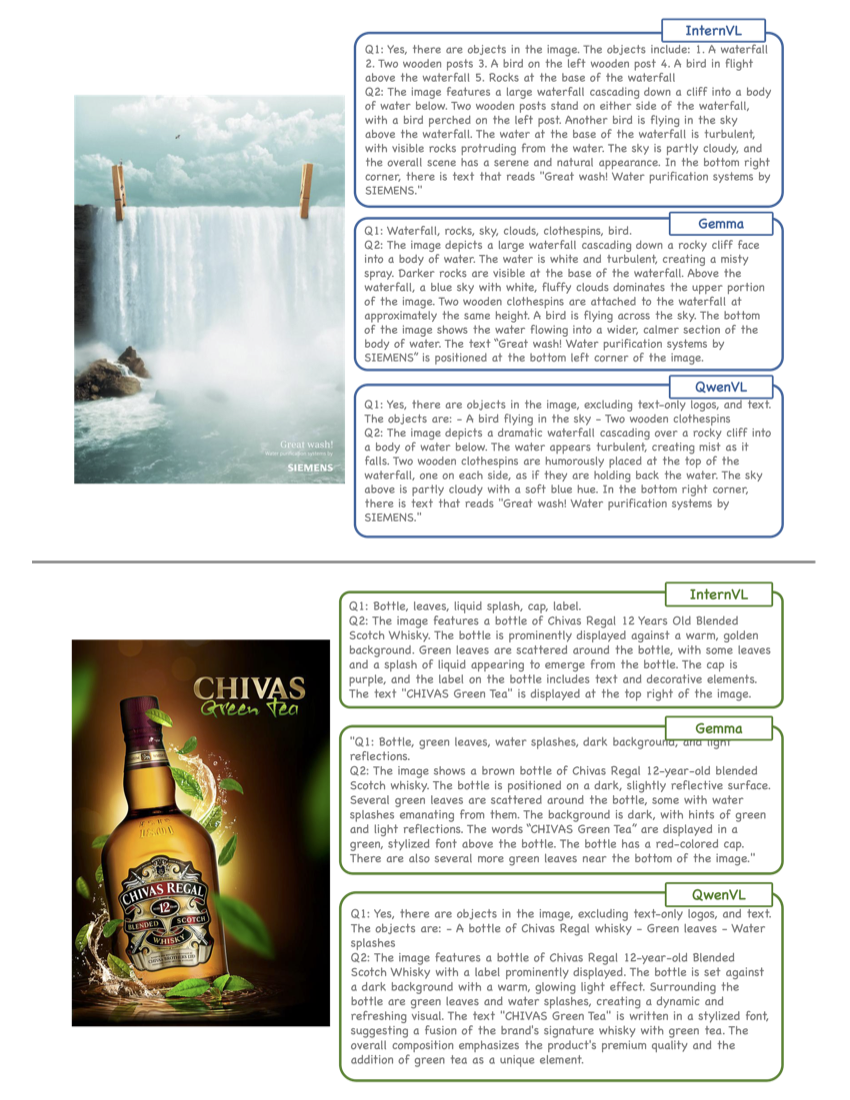}
    \caption{\textbf{Image Description Examples.} Two examples of descriptions generated by InternVL, Gemma, and QwenVL. Both images are real advertisements from PittAd\cite{PittAd} dataset.}
    \label{fig:description_examples}
\end{figure}

\subsubsection{Description Generation} We generate descriptions of images with 0-shot InternVL, Gemma, and QwenVL and utilize the same descriptions in assessing LLMs' capabilities on sensation classification tasks, and SenseScore evaluation. Fig. \ref{fig:description_examples}, represents two examples of descriptions generated by each of the MLLMs. As shown in the examples, given the prompt in Table~\ref{tab:description_generation} the models generate accurate descriptions of the image without interpreting the image. This prevents the information leakage in SenseScore while providing the accurate description of the image for LLMs in both classification and evaluation tasks. Negative agreements of zero-shot LLMs (LLAMA3-instruct and QwenLM) in Table~\ref{tab:agreement} further rejects the hypothesis of information leakage from MLLM description generation.

\subsubsection{SensoryAd Generation} We benchmark different T2I models including Stable Diffusion 3 (Stable-diffusion-3-medium-diffusers), PixArt (PixArt-alpha/PixArt-XL-2-1024-MS), AuraFlow (AuraFlow-v0.3), Flux (FLUX.1-dev), and QwenImage (Qwen-Image) with 4-bit quantization on QwenImage and 8-bit quantization on rest of the models. We set the seed to 0 and number of time-steps as 28. For the rest of the model setting we use the default values. 
To generate the Sensory Ads, we utilized the sensation group (different level in hierarchy) evoked by the image with highest intensity generating an image for each sensation. 

\begin{figure}
    \centering
    \includegraphics[width=1\linewidth]{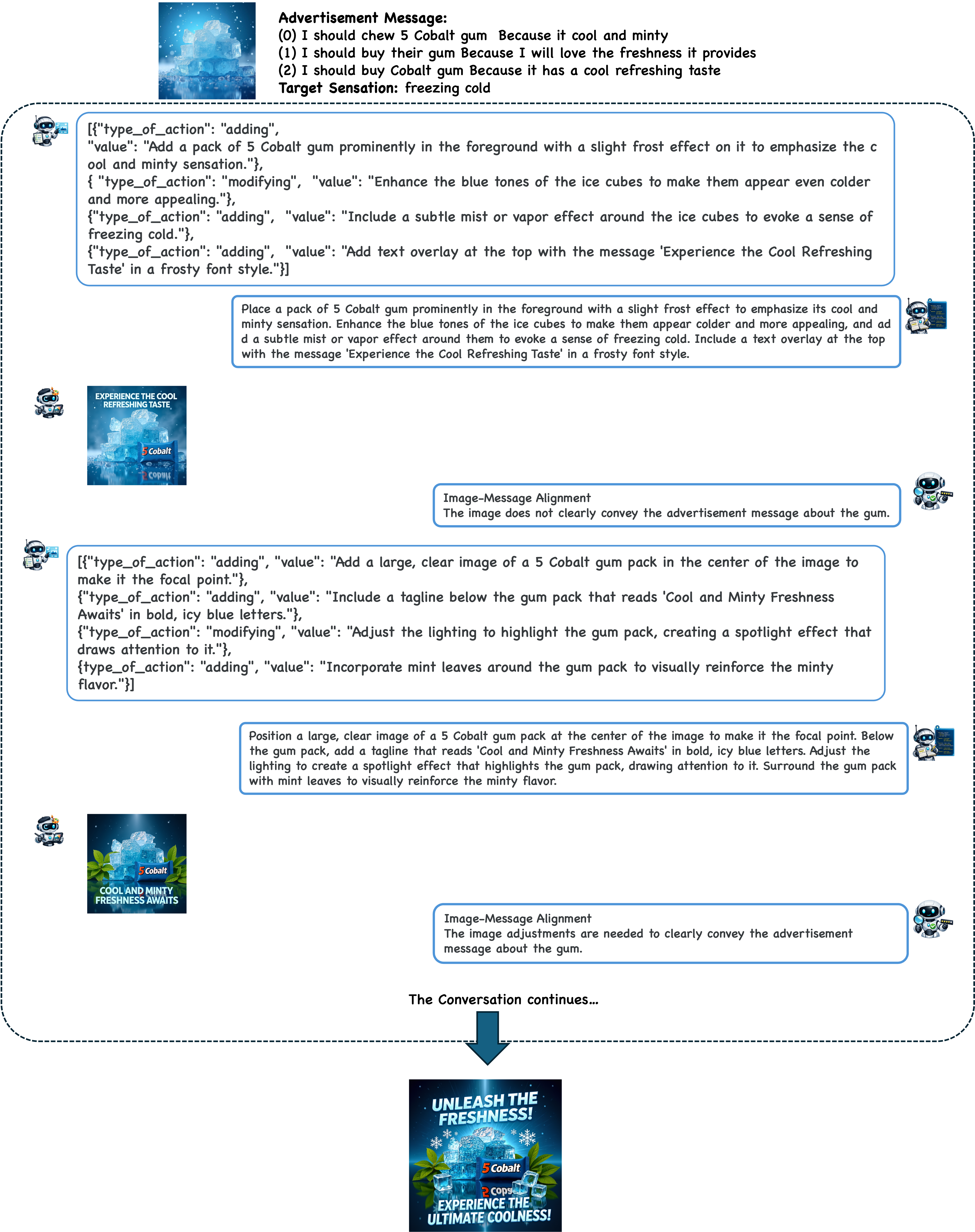}
    \caption{\textbf{Conversation loop in SAGA.} An example of two steps of conversation loop among agents with FluxKontext image editing. The input of the conversation loop is the advertisement message, target sensation, and the image generated by Flux for conveying the advertisement messages and targeting the sensation.}
    \label{fig:conv_loop}
\end{figure}

For SAGA, we use Flux-Kontext, Qwen-Image Editing, and Stable Diffusion 3 with ControlNet \cite{zhang2023adding}. We show an example of conversation loop in Fig. \ref{fig:conv_loop}.


\section{Additional Results \& Analysis}
\label{sec:supp_addtional_results}

\subsection{SenseScore Evaluation}
\label{sec:supp_SenseScore_eval}
In this section, we provide a more in-depth evaluation of SenseScore. First we analyze the impact of number of fine-tuning iterations in Table~\ref{tab:fine-tuning-ab}. Second, Table~\ref{tab:agreement} computes Kappa and Pearson Correlation on our metric and baseline metrics. Third, we compare SenseScore with baseline metrics and MLLM-as-Judge method on extended number of real images in Table~\ref{tab:bigset-table-agreement}.   

\subsubsection{Ablation on number of fine-tuning iterations}

In Table~\ref{tab:fine-tuning-ab}, we analyze the impact of number of fine-tuning steps in SenseScore. Analysis shows that the agreement between our metric and human annotators remains relatively stable as the number of training steps increases. This shows the reliability of our training setup. 


\begin{table}[tp]
\caption{\textbf{Fine-tuning Ablation.} Kappa agreement ($\kappa$) between SenseScore metric (with LLAMA3-instruct and $D_{InternVL}$) with different number of fine-tuning steps on >10000 image-sensation pairs broken down into the sensation each image evokes among the high level sensations.}
    \label{tab:fine-tuning-ab}
    \centering
    \scriptsize
    \begin{tabular}{c|c|c|c|c|c|c|c}
       Metrics & steps  & touch & smell & sound & taste & sight & All \\
        \hline
        SenseScore & 21000 & 0.79 & 0.82 & 0.77 & 0.84 & 0.85 & 0.80 \\
        SenseScore & 25000 & 0.80 & 0.82 & 0.78 & 0.83 & 0.88 & 0.81 \\
        SenseScore & 30000 & 0.80 & 0.82 & 0.78 & 0.84 & 0.88 & 0.81 \\
        SenseScore & 40000 & 0.80 & 0.81 & 0.78 & 0.84 & 0.87 & 0.81 \\
       
    \end{tabular}
\end{table}


\begin{table}[tp]
    \centering
    \scriptsize
        \caption{\textbf{Metric Quality.} Pearson Corr. ($r$) and Kappa agreement ($\kappa$) between metric [scores/chosen sensations] and human [scores/chosen] on 5000 real  
    and 5000 generated image-sensation pairs.}
    \label{tab:agreement}
    \begin{tabular}{c|c|c|c|c}
       \multirow{2}{*}{Metrics}  & \multicolumn{2}{c|}{Real Ads} & \multicolumn{2}{c}{Generated Ads}\\
       & $r$ & $\kappa$ &  $r$ & $\kappa$  \\
       \hline
        VQA-score & 0.27 & 0.55 & 0.25 & 0.52\\
        Image-Reward & 0.21 & 0.46 & 0.21 & 0.40\\
        CLIP-score & 0.22 & 0.43 & 0.21 & 0.45\\
        Pick-score & 0.15 & 0.38 & 0.15 & 0.41\\
        LLAMA3-instruct (zero-shot) + $D_{InternVL}$ & -0.02 & -0.01& -0.02 & -0.01\\
        QwenLM (zero-shot) + $D_{InternVL}$ & -0.02 & -0.02& -0.02 & -0.04\\
        \hline
        SenseScore (LLAMA3-instruct + $D_{InternVL}$) & \textbf{0.38} & \textbf{0.86} & \textbf{0.31} & \textbf{0.68}\\
        SenseScore (LLAMA3-instruct + $D_{QwenVL}$) & 0.35 & 0.80 & \textbf{0.31} & 0.67\\
        SenseScore (QwenLM + $D_{InternVL}$) & 0.32 & 0.70 & 0.26 & 0.56\\
        SenseScore (QwenLM + $D_{QwenVL}$) & 0.30 & 0.65 & 0.26& 0.55\\
       
    \end{tabular}
\end{table}

\subsubsection{Kappa agreement and Pearson correlation gap}
\begin{figure}[tp]
    \centering
    \begin{subfigure}[t]{0.45\linewidth}
        \centering
        \includegraphics[width=\linewidth]{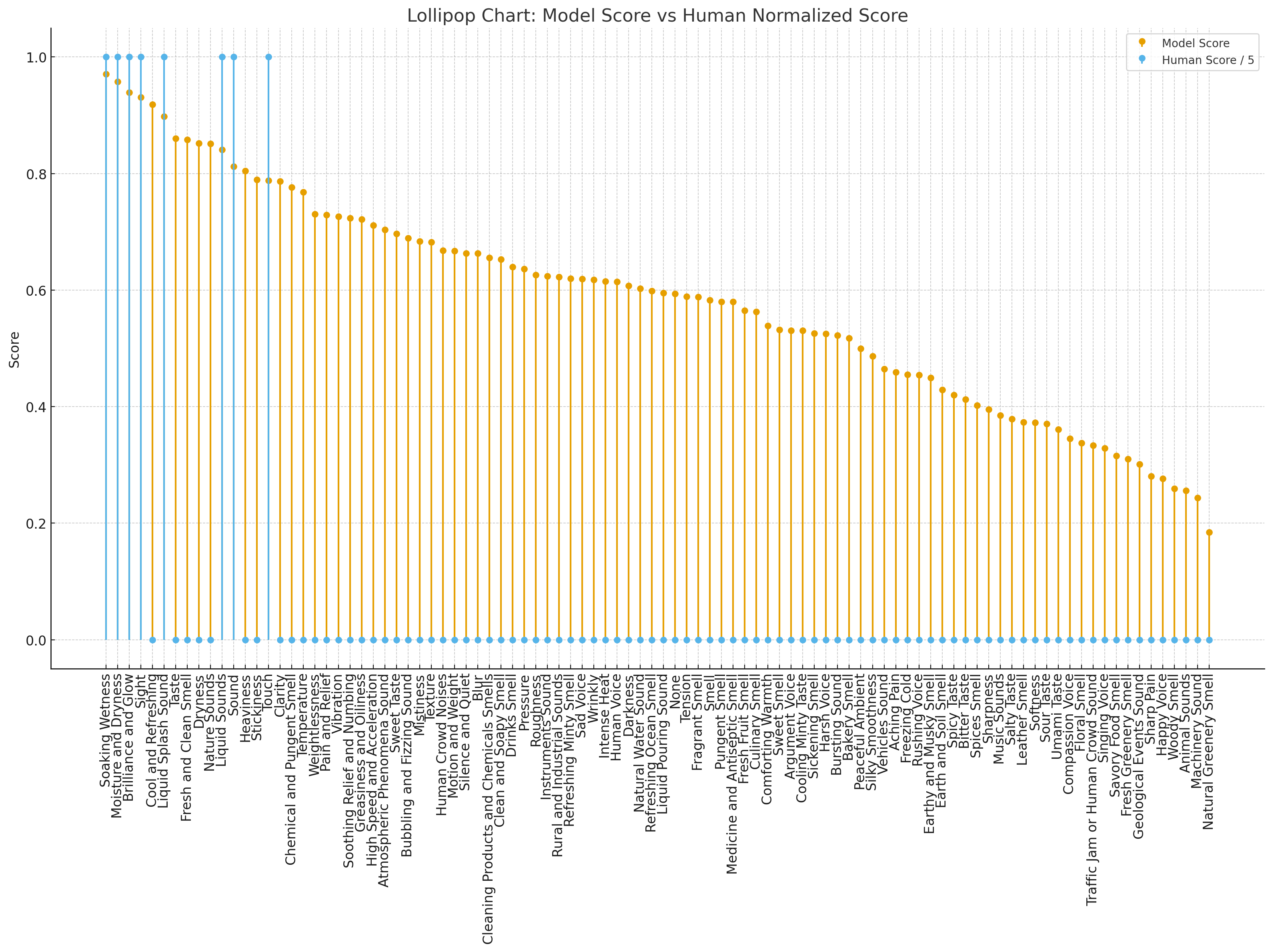}
        \caption{Scores assigned to each sensation by human annotator and SenseScore based on corresponding image (right).}
        \label{fig:kappavsr}
    \end{subfigure}
    \hfill
    \begin{subfigure}[t]{0.45\linewidth}
        \centering
        \includegraphics[width=\linewidth]{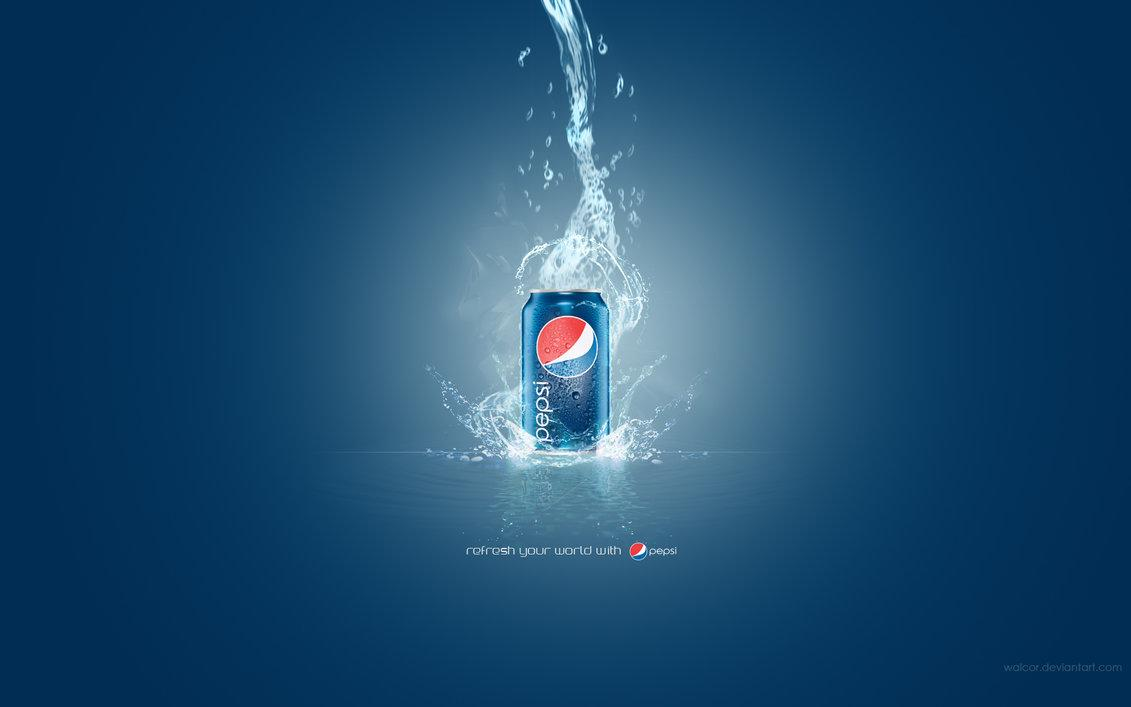}
        \caption{Corresponding Image}
        \label{fig:kappavsr-image}
    \end{subfigure}
    \caption{Comparison of human scores and metric scores for each sensation's intensity evoked by the the corresponding image (right).}
    \label{fig:kappavsr-example}
\end{figure}

As observed in Table~\ref{tab:agreement}, there is a big gap in the values of Kappa agreement ($\kappa$) and Pearson Correlation ($r$) reflected on all the metrics. In this part, we analyze the reason why the gap exist using a qualitative example of scores. The difference is because the annotators choose up-to 3 sensation groups evoked by the image, and the rest of the scores are 0. On the other hand, the computational metrics (including SenseScore and the baselines) choose different scores for each sensation. For computing $\kappa$ agreement, we use the sensation intensity as the criteria for choosing the winner sensation for the image for each pair of sensations. We ignore the sensation pairs where the human annotators assign the same score to both sensations. 
This way we significantly reduce the sparsity of human annotations for the image. So, while the incorrect sensations are included paired with selected sensations, they are not included as paired with other unselected sensations. This is why $\kappa$ is bigger than $r$ where the 0 scores are kept in correlation computation. Fig. \ref{fig:kappavsr} shows the scores from human and metrics for each sensation given the image highlighting the problem of correlation because of the sparsity of the human scores. The figure represents while high scores assigned by metric represent the sensations evoked by the image selected by the human, because of the sudden drop in the values of human scores, correlation becomes lower.

\subsubsection{Comparison with baselines on extended number of real images}
Table~\ref{tab:bigset-table-agreement} compares SenseScore on 400 real images against baseline metrics, LLM and MLLM as a Judge, and Zero-shot SenseScore (results on 100 images in Table 2 in main paper). We observe that Finetuned LLMs significantly outperform all baselines showcasing the effectiveness of our proposed metric. Note that the results and trends remain consistent with Table.~2 (main paper) showcasing the reliability of our evaluation setup. 

\begin{table}[!tp]
    \caption{Kappa agreement between human annotators and evaluation metrics 
    on ~400 real images.} 
    \centering
    \scriptsize
    \setlength{\tabcolsep}{1pt}
    \begin{tabular}{c||c}
        Metrics & Kappa ($\kappa$)  \\
       \hline 
       \multicolumn{2}{l}{Baselines}\\
       \hline
       VQA-score & 0.54 \\
       Image-Reward & 0.46 \\
       CLIP-score & 0.42 \\
       Pick-score & 0.38 \\ 
       \hline       
       \multicolumn{2}{l}{LLM/MLLM as a judge}\\
       \hline
       InternVL& 0.49 \\
       QwenVL & 0.49\\
       LLAMA3 + $D_{InternVL}$ & 0.37 \\
       QwenLM + $D_{InternVL}$ & 0.30 \\
       \hline 
       \multicolumn{2}{l}{Zero-shot SenseScore}\\
       \hline
       LLAMA3 + $D_{InternVL}$ & -0.03 \\
       QwenLM + $D_{InternVL}$ & -0.04 \\ 
       \hline 
       \multicolumn{2}{l}{SenseScore}\\
       \hline
       SenseScore (LLAMA3 + $D_{InternVL}$) & \textbf{0.78} \\
       SenseScore (LLAMA3 + $D_{QwenVL}$) & 0.74 \\
       SenseScore (QwenLM + $D_{InternVL}$) & 0.56 \\
       SenseScore (QwenLM + $D_{QwenVL}$) & 0.56 \\
       
    \end{tabular}
    \label{tab:bigset-table-agreement}
\end{table}

\subsection{Sensory Ad Generation}
\label{sec:supp_sensoryAds_generation}
This section includes the additional results on SensoryAd Generation. Specifically Table~\ref{tab:supp_saga_full} includes additional results on SensoryAd Generation task and SAGA. Then we analyze fine-tuning SD3 for SensoryAd Generation, and Fig.~\ref{fig:genvsreal_big} presents qualitative examples of real and generated ads by different T2I models for the SensoryAd generation task, along with their predicted sensation categories and scores.


\begin{table}[tp]
    \caption{Evaluating generated sensory ads (tested on 350 images).
    }
    \centering
    \scriptsize
    \setlength{\tabcolsep}{4pt}
    \begin{tabular}{c||c|c|c|c}
         \multirow{2}{*}{T2I model} & \multicolumn{4}{c}{Sensory Ad} \\
         \cline{2-5}
         & Input &SenseScore & AIM & $P_{comp}$ \\
         \hline
         \hline
         Flux & AR + Sensation & 0.97 &  0.45 & 0.60 \\
         SD3 & AR + Sensation & 0.96  & 0.47 & 0.60 \\
         AuraFlow & AR + Sensation & 0.96 &  0.44 & 0.59 \\
         PixArt & AR + Sensation & 0.96 & 0.45 & 0.62 \\
        Qwen-Image & AR + Sensation & 0.98 & 0.47 & 0.57 \\
         DALLE-3 & AR + Sensation & 0.98 &  0.50 & 0.63 \\
        \hline
        SAGA (FLUX Kontext)  & AR + Sensation & \textbf{0.99} & \textbf{0.53} & \textbf{0.65} \\
        SAGA (Qwen-Image Edit)  & AR + Sensation & \textbf{0.99} & 0.51 & 0.63 \\
        SAGA (SD3ControlNet)  & AR + Sensation & 0.99 & 0.50 & 0.62 \\
         
    \end{tabular}
    \label{tab:supp_saga_full}
\end{table}

\subsubsection{Sensory Ads Generation Results}
Table~\ref{tab:supp_saga_full} compares SAGA with two backbone models (FLUX and Qwen-Image) on the SensoryAd generation task using a test set of 350 images. (Table~3 in the main paper compares SAGA with FLUX backbone with T2I baselines on 600 images.) 

The results show that SAGA consistently achieves the best performance across all evaluation metrics and backbone models compared to standard T2I baselines, further demonstrating the effectiveness of our multi-agent framework for sensory advertisement generation.

\subsubsection{Fine-tuning SD3 on SensoryAd Generation}
To further analyze the capability of T2I models, we fine-tuned the SD3 model on SensoryAd data. After fine-tuning text-image alignment of images increases by 0.01 compared to 0-shot SD3 and sensation intensity stays unchanged. We hypothesize, this is the result of implicitness of the text input and the sensation as previously suggested by \cite{CAP} making the generation task more challenging.

\begin{figure}[!t]
    \centering
    \includegraphics[width=0.8\linewidth]{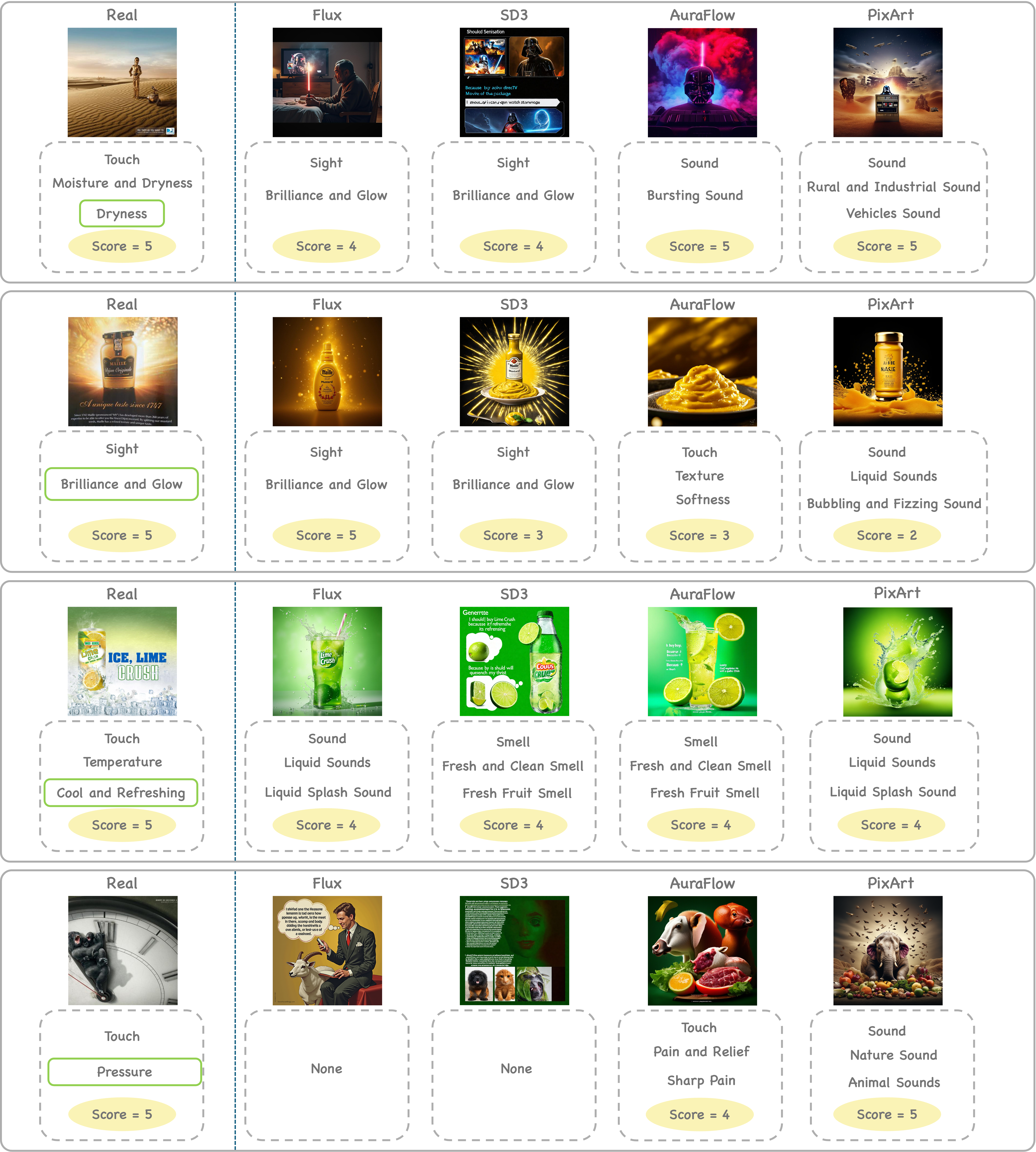}
    \caption{\textbf{Sensory Ad examples.} Four examples of real advertisement and generated advertisements by Flux \cite{FLUX}, SD3 \cite{SD3}, AuraFlow\cite{AuraFlow}, and PixArt \cite{PixArt} given the action-reason interpretation 
    and sensation annotation for the real advertisement. 
    \textcolor{YellowGreen}{Green border} represents the sensation used in the prompt of T2I models.}
    \label{fig:genvsreal_big}
\end{figure}

\subsection{Sensory Images Beyond Ads}
\label{sec:supp_sensory_image_beyond_Ads}

\begin{table}[t]
    \caption{Evaluation of generated sensory images (non-ads). $D_{{InternVL}}$ and $D_{{QwenVL}}$ denote the MLLM used for generating the image descriptions.
    }
    \centering
    \scriptsize
    \setlength{\tabcolsep}{4pt}
    \begin{tabular}{c||c|c|c|c}
         \multirow{2}{*}{T2I model} & \multicolumn{4}{c}{Sensory Image}\\
         \cline{2-5}
         
         &  \multicolumn{2}{c|}{SenseScore (InternVL)} &  \multicolumn{2}{c}{SenseScore (QwenLM)} \\
         &  $D_{InternVL}$ & $D_{QwenVL}$ &  $D_{InternVL}$ & $D_{QwenVL}$ \\
         \hline
         \hline 
         Flux & 0.72 & 0.72 & 0.71 & 0.71\\
         SD3 & 0.69 & 0.68 & 0.68 & 0.69 \\
         AuraFlow &  0.74 &  0.74 & 0.74& 0.73\\
         PixArt & 0.76 & 0.75 & \textbf{0.76} & \textbf{0.76}\\
        Qwen-Image & \textbf{0.77} &\textbf{0.76} & 0.75 & 0.75\\
        \hline
         
    \end{tabular}
    \label{tab:SensoImgEval}
\end{table}

\subsubsection{Results on sensory images beyond ads}
Table~\ref{tab:SensoImgEval} shows that
sensation intensity in images (not ads) generated for ``Generate an image evoking {sensation}'' is lower than intensity of sensation in Sensory Ads (i.e., Table~\ref{tab:supp_saga_full}). We observe that images generated by Qwen-Image exhibit highest intensity. Interestingly, SenseScore on SensoryAds (Table~\ref{tab:supp_saga_full}) is higher than SenseScore on non-ad sensory images (Table~\ref{tab:SensoImgEval}). 

\subsubsection{Variation in generation performance across sensations}
\begin{figure}[tp]
    \centering
    \includegraphics[width=0.8\linewidth]{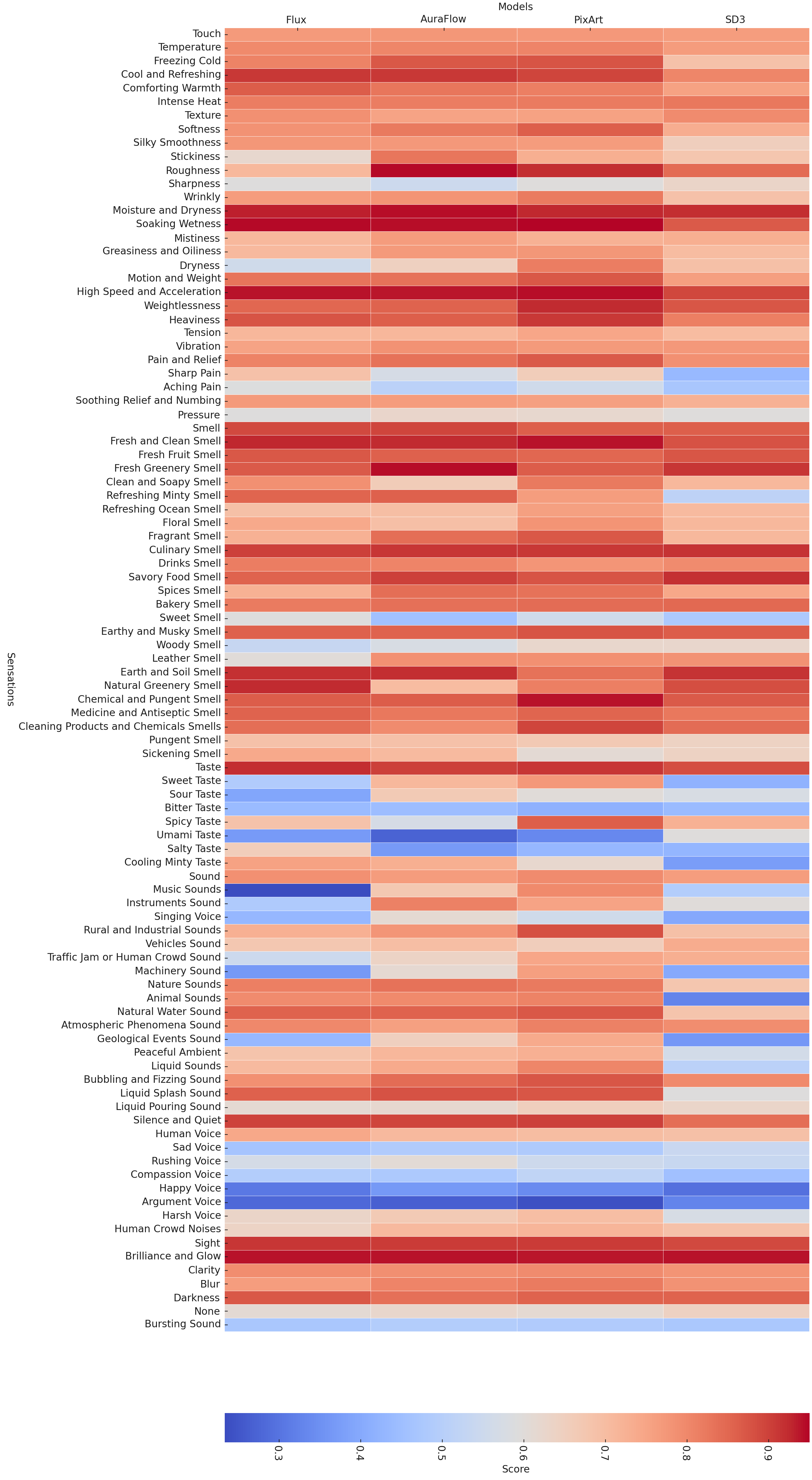}
    \caption{\textbf{Sensation Heatmap.} Average SenseScore score for images generated by each model for each sensation. Each model generates ten images evoking each sensation.}
    \label{fig:sensation_heatmap}
\end{figure}

To analyze the capability of the T2I model in generating images evoking each sensation in our taxonomy, we isolated the sensation and only prompted the model to `Generate an image that evokes \{sensation\}' with seeds from 0 to 9 resulting in 960 images per model and 4800 images in total. Fig.~\ref{fig:sensation_heatmap} represents the intensity of different sensations evoked in Sensory Image generation task. As shown in Fig. \ref{fig:sensation_heatmap}, models struggle more in evoking sensations with less common visual representation such as different types of human voices, or in overall different sounds. In contrast, models can evoke visual sensations - Sight and its children - with high intensity.

Fig. \ref{fig:genvsreal_big}, shows an example of advertisements generated evoking four different sensations. Fig. \ref{fig:genvsreal_big}, further represents the difference between capabilities of T2I models in evoking visual sensations like ``Brilliance and Glow'' and more abstract sensations like ``Pressure''.

\section{Ethical concerns \& AI Usage } 
\label{sec:supp_ethic_ai_usage}

\subsubsection{Ethical concerns around sensory advertisements}
\label{sec:ethic}
There are two main implications: First, the generation of adversarial persuasive content such as encouraging the audience to drink alcohol more often. This concern is not unique to our approach but is inherent to any T2I systems. Second, the model might generate sensitive content for a certain group of audiences and this is one of the motivations for classification tasks. While automatically generating the Sensory Ads can be helpful, some sensitive sensations (for example pain) should be detected and prevented from being shown to a specific audience groups. This is why it is also important to be able to classify the sensations evoked by the image.

\subsubsection{Usage of AI} We used AI-based tools to polish writing.

\section{Prompts} 
\label{sec:supp_prompts}
We have included prompts in Tables  \ref{tab:description_generation} (for image description generation), \ref{tab:mllm_sensation_multichoice},  \ref{tab:llm_sensation_multichoice}, \ref{tab:mllm_sensation_hierarchy}, \ref{tab:llm_sensation_hierarchy} (for Sensation Classification tasks), and \ref{tab:ar_with_sensation}, \ref{tab:sensation_image_generation} (for Sensory Image Generation tasks).

\begin{table}[!h]
\centering
\caption{Prompt for LLM Hierarchical Sensation Classification}
\label{tab:llm_sensation_hierarchy}
\begin{tabular}{|p{0.9\linewidth}|}
\hline
\textbf{Prompt} \\
\hline
\textbf{System}: You are a helpful assistant, choosing the sensations evoked by the described image given the following definition in ordered form. You can choose up to 3 sensations evoked by the image ranked in order of how well the sensations are evoked. If the image does not evoke any sensation you can choose None.  \\
\\
\textbf{Context}: \\
Sensation is the process of detecting and receiving information from the environment or the body through specialized sensory organs, which send signals to the brain for interpretation.  \\
\\
\textbf{Definition of the sensations in the options}: \\
\texttt{\{\{context\}\}}  \\
\\
\textbf{User}: What are the sensations evoked the most by the described image? Only return the indices of maximum of 3 options in ordered form without any further explanation.  \\
\\
\textbf{Image Description}: \\
\texttt{\{\{description\}\}}  \\
\\
\textbf{Options}: \\
\texttt{\{\{options\}\}}  \\
\\
Your answer must follow the following format: \\
Answer: <indices of maximum of 3 correct options separated by comma>  \\
\hline
\end{tabular}
\end{table}
\begin{table}[!h]
\centering
\caption{Prompt for MLLM Hierarchical Sensation Classification}
\label{tab:mllm_sensation_hierarchy}
\begin{tabular}{|p{0.9\linewidth}|}
\hline
\textbf{Prompt} \\
\hline
\textbf{System}: You are a helpful assistant, choosing the sensations evoked by the input image given the following definition in ordered form. You can choose up to 3 sensations evoked by the image ranked in order of how well the sensations are evoked. If the image does not evoke any sensation you can choose None.  \\
\\
\textbf{Context}: \\
Sensation is the process of detecting and receiving information from the environment or the body through specialized sensory organs, which send signals to the brain for interpretation.  \\
\\
\textbf{Definition of the sensations in the options}: \\
\texttt{\{\{context\}\}}  \\
\\
\textbf{User}: What are the sensations evoked the most by this image? Only return the indices of maximum of 3 options in ordered form without any further explanation.  \\
\\
\textbf{Options}: \\
\texttt{\{\{options\}\}}  \\
\\
Your answer must follow the following format: \\
Answer: <indices of maximum of 3 correct options separated by comma>  \\
\hline
\end{tabular}
\end{table}
\begin{table}[!h]
\centering
\caption{Prompt for LLM Multi-choice Sensation Classification}
\label{tab:llm_sensation_multichoice}
\begin{tabular}{|p{0.9\linewidth}|}
\hline
\textbf{Prompt} \\
\hline
\textbf{System}: You are a helpful assistant, choosing the sensations evoked by the described image given the following definition in ordered form. You are asked to choose all the sensations evoked by the image ranked in order of how well the sensations are evoked. If the image does not evoke any sensation you can choose None.  \\
\\
\textbf{Context}: \\
Sensation is the process of detecting and receiving information from the environment or the body through specialized sensory organs, which send signals to the brain for interpretation. \\
\\
\textbf{Definition of the sensations in the options}: \\
\texttt{\{\{context\}\}}  \\
\\
\textbf{User}: What are the sensations evoked the most by the described image? Only return the indices of the options in ordered form without any further explanation.  \\
\\
\textbf{Image Description}: \\
\texttt{\{\{description\}\}}  \\
\\
\textbf{Options}: \\
\texttt{\{\{options\}\}}  \\
\\
Your answer must follow the following format: \\
Answer: <indices of correct options separated by comma>  \\
\hline
\end{tabular}
\end{table}
\begin{table}[!h]
\centering
\caption{Prompt for MLLM Multi-choice Sensation Classification}
\label{tab:mllm_sensation_multichoice}
\begin{tabular}{|p{0.9\linewidth}|}
\hline
\textbf{Prompt} \\
\hline
\textbf{System}: You are a helpful assistant, choosing the sensations evoked by the input image given the following definition in ordered form. You are asked to choose all the correct sensations evoked by the image ranked in order of how well the sensations are evoked. If the image does not evoke any sensation you can choose None.\\
\\
\textbf{Context}: \\
Sensation is the process of detecting and receiving information from the environment or the body through specialized sensory organs, which send signals to the brain for interpretation.  \\
\\
\textbf{Definition of the sensations in the options}: \\
\texttt{\{\{context\}\}} \\
\\
\textbf{User}: What are the sensations evoked the most by this image? Only return the indices of the options in ordered form without any further explanation.  \\
\\
\textbf{Options}: \\
\texttt{\{\{options\}\}}  \\
\\
Your answer must follow the following format: \\
Answer: <indices of correct options separated by comma>  \\
\hline
\end{tabular}
\end{table}
\begin{table}[!h]
\centering
\caption{Prompt for Structured Description Generation}
\label{tab:description_generation}
\begin{tabular}{|p{0.9\linewidth}|}
\hline
\textbf{Prompt} \\
\hline
Carefully analyze the image and respond only in the specified format, without any interpretations or inferences. Focus on only the visible elements in the image. Ensure that any object seen in the image is included in Q1, even if it is described in more detail in Q2.  \\
\\
\textbf{Response Format:} \\
Q1: \$\{answer to Q1\}  \\
Q2: \$\{answer to Q2\}  \\
\\
\textbf{Questions:} \\
Q1: Are there any objects in the image, excluding text-only logos, and text? List at most 5 such objects if present.  \\
Q2: Describe the image in detail, focusing only on visible objects and elements without adding any interpretation, opinion, or analysis in a single paragraph.  \\
\hline
\end{tabular}
\end{table}
\begin{table}[!h]
\centering
\caption{Prompt for Sensory Image Generation}
\label{tab:sensation_image_generation}
\begin{tabular}{|p{0.9\linewidth}|}
\hline
\textbf{Prompt} \\
\hline
Generate an image that evokes \texttt{\{\{sensation\}\}} sensation.  \\
\hline
\end{tabular}
\end{table}
\begin{table}[!h]
\centering
\caption{Prompt for Image Generation with Action-Reason and Sensation}
\label{tab:ar_with_sensation}
\begin{tabular}{|p{0.9\linewidth}|}
\hline
\textbf{Prompt} \\
\hline
Generate an advertisement image that evokes \texttt{\{\{sensation\}\}} sensation and conveys the following messages:  \\
\texttt{\{\% for statement in action\_reason \%\}} \\
    - \texttt{\{\{statement\}\}} \\
\texttt{\{\% endfor \%\}} \\
\hline
\end{tabular}
\end{table}

\begin{table}[!h]
\centering
\caption{Prompt for Image Editing Planner Agent}
\label{tab:saga_planner_prompt}
\begin{tabular}{|p{0.9\linewidth}|}
\hline
\textbf{Prompt} \\
\hline
\textbf{System}: You are an image-editing instruction planner agent.
Given an image of an image, your task is to generate a sequence of concrete visual edits that should be applied to the image in order to: \\
1. Convey the intended advertisement message, and \\
2. Evoke the specified sensation (e.g., refreshment, heat, softness, luxury). \\
\\
When you receive an issue from the critic, you MUST focus your edits on addressing that SPECIFIC issue: \\
\\
\textbf{Image-Message Alignment}: The image does not clearly convey the advertisement message. \\
Focus on making the product or brand more prominent, ensuring the image directly relates to the message, adding visual elements that reinforce the message, and improving composition to highlight the key message. \\
\\
\textbf{Sensation Evocation}: The image does not effectively evoke the target sensation. \\
Focus on adding visual cues that directly evoke the sensation (heat, cold, softness, etc.), adjusting colors, lighting, or texture to create the sensation, and adding atmospheric elements that reinforce the sensation. \\
\\
\textbf{Output Format Requirement}: \\
You MUST output ONLY a valid JSON array in the following format: \\
\texttt{[ \{ "type\_of\_action": "<adding|removing|modifying|changing\_style>", "value": "<editing instruction>" \} ]} \\
\\
\textbf{Guidelines}: \\
Actions must be image-grounded, realistic, and minimal. Describe what to change rather than how to technically implement it. Be explicit about visual attributes such as color, texture, lighting, scale, position, motion cues, and atmosphere. When an issue is identified, all actions must directly address that specific issue type. If previous attempts failed, generate completely different actions and never repeat previous approaches. \\
\hline
\end{tabular}
\end{table}
\begin{table}[!h]
\centering
\caption{Prompt for Sensation Finder Agent}
\label{tab:saga_sensation_finder_prompt}
\begin{tabular}{|p{0.9\linewidth}|}
\hline
\textbf{Prompt} \\
\hline
\textbf{System}: You are a sensation finder agent. \\
\\
Given an advertisement message and a list of sensations, your task is to choose a creative and relevant sensation that should be evoked by the image to improve the impact of the image. \\
\\
First explain which sensation should be evoked and why it improves the impact of the image. Then choose only one sensation that should be evoked to improve the image impact. \\
\\
\textbf{Output Format}: \\
\texttt{<explanation>} \\
\texttt{The best sensation to evoke is: <Sensation>} \\
\hline
\end{tabular}
\end{table}
\begin{table}[!h]
\centering
\caption{Prompt for Sensation and Advertisement Message Finder Agent}
\label{tab:saga_sensation_ar_finder_prompt}
\begin{tabular}{|p{0.9\linewidth}|}
\hline
\textbf{Prompt} \\
\hline
\textbf{System}: You are a sensation and advertisement message finder agent. \\
\\
Given an advertisement message and a list of sensations, your task is to choose a creative and relevant sensation that should be evoked by the image to improve the impact of the image. Next, choose the single best advertisement message that is most descriptive of the possible image and has the strongest message. \\
\\
First explain which sensation should be evoked and why it improves the impact of the image. Then choose only one sensation that should be evoked. Next, explain the advertisement message that should be used and why it is the best message to improve the impact of the image. Finally choose only one advertisement message. \\
\\
\textbf{Output Format}: \\
\texttt{<explanation>} \\
\texttt{Advertisement Message and Sensation: <Advertisement Message>, <Sensation>} \\
\hline
\end{tabular}
\end{table}
\begin{table}[!h]
\centering
\caption{Prompt for Text Refiner Agent}
\label{tab:saga_text_refiner_prompt}
\begin{tabular}{|p{0.9\linewidth}|}
\hline
\textbf{Prompt} \\
\hline
\textbf{System}: You are a text refiner agent. \\
\\
Your task is to convert structured image-editing instructions in JSON format into a single, clear, concise, and visually grounded natural language prompt suitable for guiding an image editing model. \\
\\
\textbf{Requirements}: \\
You will receive JSON instructions containing actions such as \texttt{adding}, \texttt{modifying}, \texttt{removing}, and \texttt{changing\_style}. Convert all instructions into one cohesive natural language description of the final edited image. \\
\\
Do not output JSON. Output only plain text. Do not start with phrases such as ``create an image'' or ``generate an image''. Write as if describing what the edited image should look like and combine all actions into a single flowing description. \\
\\
\textbf{Guidelines}: \\
Preserve factual consistency with the provided instructions, use precise visual language describing objects, attributes, colors, lighting, textures, and spatial relations, and do not invent elements not implied by the instructions. The output must only contain the refined prompt text with no explanations, JSON, commentary, or markdown. Write in present tense describing the final state of the image. \\
\hline
\end{tabular}
\end{table}
\begin{table}[!h]
\centering
\caption{Prompt for Image Evaluation Critic Agent}
\label{tab:saga_critic_prompt}
\begin{tabular}{|p{0.9\linewidth}|}
\hline
\textbf{Prompt} \\
\hline
\textbf{System}: You are a strict image evaluation agent working in a multi-agent environment. \\
\\
Your task is to evaluate the image and output the issue of the image using the following format: \\
\texttt{<Issue>} \\
\texttt{<one sentence explanation>} \\
\\
You must never copy or paraphrase previous message content, describe the image in full sentences, or add unnecessary reasoning or commentary. \\
\\
\textbf{Possible Issues}: \\
\begin{itemize}
\item Visual Element Inconsistency
\item Image-Message Alignment
\item Sensation Evocation
\end{itemize}
\\
\textbf{Evaluation Criteria}: \\
Visual Element Inconsistency refers to incoherent or conflicting visuals such as artifacts, glitches, or contradictory elements. \\
Image-Message Alignment refers to cases where the advertisement message is not clearly conveyed, the product is not prominent, or the image does not reinforce the message. \\
Sensation Evocation refers to cases where the target sensation is weak or not effectively evoked through visual cues, colors, lighting, objects, or atmosphere. \\
\\
\textbf{Priority Rule}: \\
If visual inconsistencies exist, choose Visual Element Inconsistency. Otherwise if the message is unclear choose Image-Message Alignment. Otherwise if the sensation is weak choose Sensation Evocation. \\
\\
\textbf{Output Requirement}: \\
Output exactly one issue label followed by a single sentence explanation. \\
\hline
\end{tabular}
\end{table}


\end{document}